\documentclass{article}
\usepackage{nips10submit_e,times}
\usepackage{amssymb, amsmath, amsthm,subfigure}
\usepackage{graphicx}
\usepackage[breaklinks=true,letterpaper=true,colorlinks,citecolor=blue,bookmarks=false]{hyperref}
\usepackage[numbers,sort,compress,square,comma]{natbib}
\newtheorem{thm}{Theorem}[section]

\newtheorem{prop}[thm]{Proposition}

\newcommand{\ignore}[1]{\textbf{}}

\newcommand{\stable}[1]{\S(\alpha,\beta_{#1},\gamma_{#1},\delta_{#1})}

\newcommand{\diag}{\mathop{\bf diag}}

\newcommand{\conv}[1]{\mathop{\stackrel{*}{\prod}} \limits_{#1}}

\def\BE{\begin{equation}}
\def\BES{\small  \[ }
\def\EES{\] \vspace{-2mm} \normalsize}
\def\EE{\end{equation}}
\def\BEA{\begin{eqnarray}}
\def\EEA{\end{eqnarray}}
\newcommand{\cut}[1]{{}}

\newcommand\vx{{x}}
\newcommand\vy{{y}}
\newcommand\vz{{z}}
\newcommand\mA{{A}} 

\newcommand\mD{{\bf D}}

\newcommand\mR{{R}}

\newcommand\F{\mathcal{F}}
\newcommand\R{\mathbb{R}}

\newcommand\N{\mathcal{N}}

\renewcommand\S{\mathcal{S}}

\newtheorem{theorem}{Theorem}[section]

\newtheorem{defs}[theorem]{Definition}

\DeclareMathOperator*{\sign}{sign}

\title{Inference with Multivariate Heavy-Tails\\ in Linear Models}

\author{
Danny Bickson and Carlos Guestrin \\
Machine Learning Department\\
Carnegie Mellon University\\
Pittsburgh, PA 15213 \\
\texttt{\{bickson,guestrin\}@cs.cmu.edu} \\
}

\begin{document}

\nipsfinalcopy
\vspace{-2mm}
\maketitle
\vspace{-2mm}
\begin{abstract}
\vspace{-2mm}
{\bf } Heavy-tailed distributions naturally occur in many real life problems. Unfortunately, it is typically not possible to compute inference in closed-form in graphical models which involve such heavy-tailed distributions. 

In this work, we propose a novel simple linear graphical model for  independent latent random variables, called linear characteristic model (LCM), defined in the characteristic function domain. Using stable distributions, a heavy-tailed family of distributions which is a generalization of Cauchy, L\'evy and Gaussian distributions, we show for the first time, how to compute both exact and approximate inference in such a linear multivariate graphical model.
LCMs are not limited to  stable distributions, in fact LCMs are always defined for any random variables (discrete, continuous or a mixture of both). 

We provide a realistic problem from the field of computer networks to demonstrate the applicability of our construction. 
Other potential application is iterative decoding of linear channels with non-Gaussian noise.    \end{abstract}
\vspace{-2mm}
\section{Introduction}
Heavy-tailed distributions naturally occur in many real life phenomena, for example  in computer networks \cite{NetLoss,Flows4,Flows3}. Typically, a small set of machines are responsible for a large fraction of the consumed network bandwidth. Equivalently, a small set of users generate a large fraction of the network traffic.  Another common property of communication networks is that network traffic tends to be linear \cite{ATANDT,NetLoss}. Linearity is explained by the fact that the total incoming traffic at a node is composed from the sum of distinct incoming flows. 

Recently, several works propose to use linear multivariate statistical methods for monitoring network health, performance analysis or intrusion detection \cite{Flows1,Flows2,Flows3,Flows4}.
Some of the aspects of network traffic makes the task of modeling it using
a probabilistic graphical models challenging. In many cases, the underlying heavy-tailed distributions are difficult to work with analytically. That is why existing solutions in the area of network monitoring involve various approximations of the joint probability distribution function using a variety of techniques:\ mixtures of distributions \cite{ATANDT}, spectral decomposition \cite{Flows2} historgrams \cite{Flows4}, sketches \cite{Flows3}, entropy \cite{Flows4}, sampled moments \cite{NetLoss}, etc.

In the current work, we propose a novel linear probabilistic graphical model called linear characteristic model (LCM) to model linear interactions of independent heavy-tailed random variables (Section 3). Using the stable family of distributions (defined in Section 2), a family of heavy-tailed distributions, we show how to compute both exact and approximate inference (Section 4). Using real data from the domain of computer networks we demonstrate the applicability of our proposed methods for computing inference in LCM (Section 5).

We summarize our  contributions below:
\begin{itemize}

\item We propose a new linear graphical model called LCM, defined as a product of factors in the cf\ domain.  We show that our model is well defined for any collection of random  variables, since any random variable has a matching cf.

\item Computing inference in closed form in linear models involving continuous variables is typically  limited to the well understood cases
of Gaussians  and simple regression problems in exponential families. In this work, we extend the applicability of belief propagation to the stable family of distributions, a generalization of Gaussian, Cauchy and L\'evy distributions. We analyze both exact and approximate inference algorithms, including convergence and accuracy of the solution.
\item We demonstrate the applicability of our proposed method, performing inference in real settings, using network tomography data obtained from the PlanetLab network.\end{itemize}
\vspace{-2mm}
\subsection{Related work}
There are three main relevant works in the machine learning domain which are related to the current work:\ Convolutional Factor Graphs (CFG), Copulas and Independent Component Analysis (ICA). Below we shortly review them and motivate why a new graphical model is needed. 

Convolutional Factor Graphs (CFG) \cite{CFG0,CFG} are a graphical model for representing  linear relation of independent latent random variables. CFG assume that the probability distribution factorizes as a convolution of potentials, and proposes to use duality to derive a product factorization in the characteristic function (cf) domain. In this work we extend CFG by defining the graphical model as a product of factors in the cf domain. Unlike CFGs, LCMs are always defined, for any probability distribution, while CFG may are not defined when the inverse Fourier transform does not exist.  

A closely related technique is the Copula method \cite{Copula,Nonparanormal}.
Similar to our work, Copulas assume a linear underlying model. The main difference is that Copulas  transform each marginal variable into a uniform distribution and perform inference in the cumulative distribution function (cdf)\ domain. In contrast, we perform inference in the cf domain.  In our case of interest, when the underlying distributions are stable, Copulas can not be used since stable distributions are not analytically expressible in the cdf domain.

A third  related technique is ICA (independent component analysis) on linear models \cite{ICA}. Assuming a linear model $Y=AX$\footnote{Linear model is formally defined in Section 3.}, where the observations $Y$ are given, the task is to estimate the linear relation matrix A, using only the
fact that the latent variables X are statistically mutually
independent. Both techniques (LCM and ICA)\ are complementary, since ICA can be used to learn the linear model, while LCM is used for computing inference in the learned model.






\vspace{-2mm}
\section{Stable distribution}
Stable distribution \cite{Zoltarev} is a family of heavy-tailed distributions, where Cauchy, L\'evy and Gaussian are special instances of this family (see Figure 1). Stable distributions are used in different problem domains, including economics, physics, geology and astronomy \cite{Nolan2}. Stable distribution are useful since they can model  heavy-tailed distributions that naturally occur in practice. As we will soon show with our networking example, network flows exhibit empirical distribution which can be modeled remarkably well by stable distributions. 

We denote a stable distribution by a tuple of four parameters:  $\S(\alpha,\beta,\gamma,\delta)$. We call $\alpha$ as the characteristic exponent, $\beta$ is the skew parameter, $\gamma$ is a scale parameter and $\delta$ is a shift parameter.  For example (Fig.~1), a Gaussian $\N(\mu,\sigma^2)$ is a stable distribution with the parameters  $\S(2,0,\tfrac{\sigma}{\sqrt{2}},\mu)$, a Cauchy distribution $\mathtt{Cauchy}(\gamma,\delta)$ is stable with $\S(1,0,\gamma,\delta)$ and a L\'evy distribution $\mathtt{\mbox{L\'evy}}(\gamma,\delta)$ is stable with $\S(\tfrac{1}{2},1,\gamma,\delta)$.
Following we define formally a stable distribution. We begin by  defining a unit scale, zero-centered stable random variable.

\begin{defs}\cite[Def. 1.6]{NOLAN}
A random variable $X$ is {\em stable} if and only if $X\sim  aZ+b$,   $0 < \alpha \le 2$, $-1 \le \beta \le 1$,  $a,b \in \R$, $a \ne 0$ and $Z$ is a random variable with characteristic function\footnote{We formally define characteristic function in the supplementary material.} \small \BE E[\exp(iuZ)]=  \begin{cases}\exp\big(-|u|^\alpha[1-i \beta \tan(\tfrac{\pi \alpha}{2})\sign(u)]\big)& \alpha \ne 1\\
\exp\big(-|u|[1+i \beta \tfrac{2}{\pi}\sign(u)\log(|u|)]\big) & \alpha = 1 \end{cases}\,. \label{char_func}
\EE \normalsize
\end{defs}
Next we define a general stable random variable.
 \begin{defs}\cite[Def. 1.7]{NOLAN} 
 A random variable $X$ is $\S(\alpha,\beta,\gamma,\delta)$ if \small \[ X \sim \begin{cases} \gamma(Z-\beta \tan(\tfrac{\pi \alpha}{2}))+\delta & \alpha \ne 1 \\ \gamma Z + \delta & \alpha = 1 \end{cases} \,,\] \normalsize
 where $Z$ is given by \eqref{char_func}. $X$ has characteristic function
 \small \[ E\exp(iuZ)=  \begin{cases}\exp(-\gamma^\alpha|u|^\alpha[1-i \beta \tan(\tfrac{\pi \alpha}{2}) \sign(u)(|\gamma u|^{1-\alpha}-1)]+i \delta u )& \alpha \ne 1\\
 \exp(-\gamma|u|[1+i \beta \tfrac{2}{\pi}\sign(u)\log(\gamma|u|)]+i \delta u) &\alpha = 1 \end{cases} \label{char_mult}\,. \] \normalsize
 \end{defs}\vspace{-2mm}
A basic property of stable laws is that weighted sums of $\alpha$-stable random variables is $\alpha$-stable (and hence the family is called stable). This property will be useful  in the next section where we compute inference in a linear graphical model with underlying stable distributions. The following proposition formulates this linearity.
\begin{prop}\cite[Prop. 1.16]{NOLAN}\label{prop_s2}\begin{enumerate}
\item[a) ]{\bf Multiplication by a scalar.} If $X \sim \S(\alpha,\beta,\gamma,\delta)$ then for any $a,b \in \R, a\ne 0\,,$  \\\vspace{-2mm}
\small \[ aX+b \sim \S(\alpha,\sign(a)\beta,|a|\gamma,a\delta+b)\,.\] \normalsize
\item[b) ]{\bf Summation of two stable variables.} If $X_1 \sim \S(\alpha,\beta_1, \gamma_1,\delta_1)$ and 
 $X_2 \sim \S(\alpha,\beta_2, \gamma_2,\delta_2)$ are independent, then  $X_1+X_2 \sim \S(\alpha,\beta,\gamma,\delta)$ where
\[ \beta = \frac{\beta_1\gamma_1^\alpha+\beta_2\gamma_2^\alpha}{\gamma_1^\alpha+\gamma_2^\alpha}, \ \ \ \gamma^\alpha=\gamma_1^\alpha+\gamma_2^\alpha\ \ ,\ \ \ \  \delta=\delta_1+\delta_2+\xi\,,\]
 \[ \!\!\!\!\!\!\!\!\!\!\!\!\!\! \xi=\begin{cases} \tan(\tfrac{\pi \alpha}{2})[\beta\gamma-\beta_1\gamma_1-\beta_2\gamma_2]& \alpha \ne 1\\
\tfrac{2}{\pi}[\beta\gamma\log\gamma-\beta_1\gamma_1\log\gamma_1-\beta_2\gamma_2\log\gamma_2] & \alpha = 1\end{cases}\,. \]
Note that both $X_1,X_2$ have to be distributed with the same characteristic exponent $\alpha$.
\end{enumerate}
\end{prop} \vspace{-2mm}
\vspace{-2mm}
\section{Linear characteristic models}
A drawback of general stable distributions, is that they do not have closed-form equation for the pdf or the cdf. This fact makes the handling of stable distributions more difficult. This is probably one of the reasons stable distribution are  rarely used in the probabilistic graphical models community. 

We propose a novel approach for modeling  linear interactions between random variables distributed according to stable distributions, using a new linear probabilistic graphical model called LCM. A new graphical model is  needed, since previous approaches like CFG or  the Copula method can not be used for computing inference in closed-form in linear models involving stable distribution, because they require computation in the pdf or cdf domains respectively.
We start by defining a linear model:
\begin{defs}(Linear model) \label{linear}
Let $X_1,\cdots,X_n$  a set of mutually independent  random variables.\footnote{We do not limit the type of random variables. The variables may be discrete, continuous, or a mixture of both. } Let $Y_1,\cdots,Y_m$ be a set of observations obtained using the linear model:
\small
\[ Y_i \sim \sum_{j}A_{ij}X_{j} \ \ \ \ \forall_i\,, \]
\normalsize
where $A_{ij} \in \mathbb{R}$ are weighting scalars.
We denote the linear model in matrix notation as $Y=AX$.\end{defs}
Linear models are  useful in many domains. For example, in linear channel decoding, $X$ are the transmitted codewords, the matrix $A$ is the linear channel transformation and $Y$ is a vector of observations. When $X$ are distributed using a Gaussian distribution, the channel model is called AWGN (additive white Gaussian noise) channel. Typically, the decoding task is finding the most probable $X,$ given $A$ and the observation $Y.$ Despite the fact that $X$ are assumed statistically mutually independent when transmitting, given an observation   $Y$, $X$ are not independent any more, since they are correlated via the observation. Besides of the network application we focus on, other potential application to our current work is linear channel decoding with stable, non-Gaussian, noise.

In the rest of this section we develop the foundations for computing inference in a linear model using underlying stable distributions.
Because stable distributions do not have closed-form equations in the pdf domain, we must work in the cf domain. Hence, we define a dual linear model in the cf domain. 



 
\subsection{Duality of LCM and CFG}
CFG  \cite{CFG} have shown that the joint probability $p(x,y)$ of any linear model can be factorized as a convolution:\small\BE p(x,y) =p(x_1,\cdots,x_n,y_1,\cdots,y_m)=\conv{i}p(x_i,y_1,\cdots,y_m)\,. \label{CFG_model} \EE  \normalsize  
 Informally, LCM is the dual representation of \eqref{CFG_model} in the characteristic function domain. Next, we define LCM formally, and  establish the duality to the factorization given in \eqref{CFG_model}. 
\begin{defs}(LCM) \label{lcmd}Given the linear model Y=AX, we define the linear characteristic model (LCM)\small
\[ \varphi(t_1,\cdots,t_n,s_1,\cdots,s_m)\triangleq \prod_i \varphi(t_i,s_1,\cdots,s_m)\,,\]\normalsize
where $\varphi(t_i,s_1,\cdots,s_m)$ is the characteristic function\footnote{Defined in the supplementary material.} of the joint distribution $p(x_i,y_1,\cdots,y_m)$.
\end{defs}

\normalsize
 
The following two theorems establish duality between the LCM  and its dual representation in the pdf domain.
This duality is well known (see for example \cite{CFG0,CFG}), but important for explaining the derivation of LCM from the linear model.  \begin{theorem}\label{duality}  Given a LCM, assuming $p(x,y)$ as defined in (2) has a closed form and the Fourier transform $\mathcal{F}[p(x,y)]$ exists, then the  $\mathcal{F}[p(x,y)] =\varphi(t_1,\cdots,t_n,s_1,\cdots,s_m)$.
\end{theorem}

\begin{theorem}\label{duality2}
Given a LCM,~when the inverse Fourier transform exists, then $\F^{-1}(\varphi(t_1,\cdots,t_n,s_1,\cdots,s_m)) =  p(x,y)$\ as defined in (2). 
\label{lcm2marginal}
\end{theorem}

The proof of all theorems is deferred to  the supplementary material.
Whenever the inverse Fourier transform exists, LCM model has a dual CFG model. In contrast to the CFG model, LCM are always defined, even the inverse Fourier transform does not exist. The duality is useful, since it allows us to compute inference in either representations, whenever it is more convenient. 
\vspace{-2mm}
\section{Main result:  exact and approximate inference in LCM}
This section brings our main result. Typically, exact inference in linear models with continuous variables is limited to the well understood cases of Gaussian and simple regression problem in exponential families. In this section we extend previous results, to show how to compute inference (both exact and approximate)\ in linear model with underlying stable distributions. 
%
\vspace{-2mm}\subsection{Exact inference in LCM}\vspace{-2mm}
The inference task typically involves computation of marginal distribution or a conditional distribution of a probability function. For the rest of the discussion we focus on marginal distribution. Marginal distribution of the node $x_i$ is typically computed by integrating out all other nodes:

\small \[ p(x_i|y) \sim \int \limits_{X\setminus i } p(x,y)\ d_{X{\setminus i}}\,, \] \normalsize
where \small\ $X\setminus i $\normalsize \  is the set of all nodes excluding node $i$.    Unfortunately, when working with stable distribution, the above integral is intractable. Instead, we  propose to use a dual operation called slicing, computed in the cf domain.   

\begin{defs}(slicing/evaluation)\cite[p. 110]{Soong}

(a) {\bf Joint cf}. Given random variables $X_1, X_2$, the {\em joint cf} is
$ \varphi_{X_1, X_2}(t_1,t_2)= E[e^{it_1x_1+it_2x_2}] $.\\
(b) {\bf Marginal cf}. The {\em marginal cf} is derived  from the joint cf by $\varphi_{X_1}(t_1) = \varphi_{X_1,X_2}(t_1, 0).$ This operation is called slicing or evaluation.
We denote the slicing operation as \small$\varphi_{X_1}(t_1)=\varphi_{X_1,X_2}(t_1,t_2) \bigg]_{t_2=0}
$. \normalsize\vspace{-2mm}
\end{defs}
The following theorem establishes the fact that marginal distribution can be computed in the cf domain, by using the slicing operation. \begin{theorem}\label{lcm3}
Given a LCM, the marginal\ cf of the random variable $X_i$ can be computed using
\small
\BE \varphi_{}(t_i) = \prod_j \varphi(t_{j},s_{1},\cdots,s_m)\bigg]_{{T \setminus i}=0} \, \label{marginal}\,, \EE
\normalsize
 In case the inverse Fourier transform exists, then the marginal probability
of the hidden variable $X_i$ is given by \small$p(x_i) \sim \F^{-1}\{\varphi(t_i)\}\,.$\normalsize
\end{theorem}
\begin{figure}[t!]
 \fbox{
 \begin{minipage}[t!]{2.5in}
\scriptsize
 {\tt for }$i \in  |T|$\\
 \{\\
 {\tt Eliminate $t_i$ by computing}
 \vspace{-2mm}
 \[ \phi_{m+i}(N(t_i)) = \prod_{\varphi_j \in N(t_i)} \phi(t_j,s_1,\cdots,s_m)\bigg]_{t_i=0}  \]
{\tt Remove $\phi(t_j,s_1,\cdots,s_m)$ and $t_i$ from LCM. 

Add $\phi_{m+i} $ to LCM. \\}
 \}\\
 {\tt Finally: If $\F^{-1}$ exists, compute \[p(x_{i})=\F^{-1}(\phi_{final})\,.\] }
\end{minipage}
}
\begin{minipage}[t!]{2.5in}
\includegraphics[scale=0.30,clip,bb=28 180 600 600]{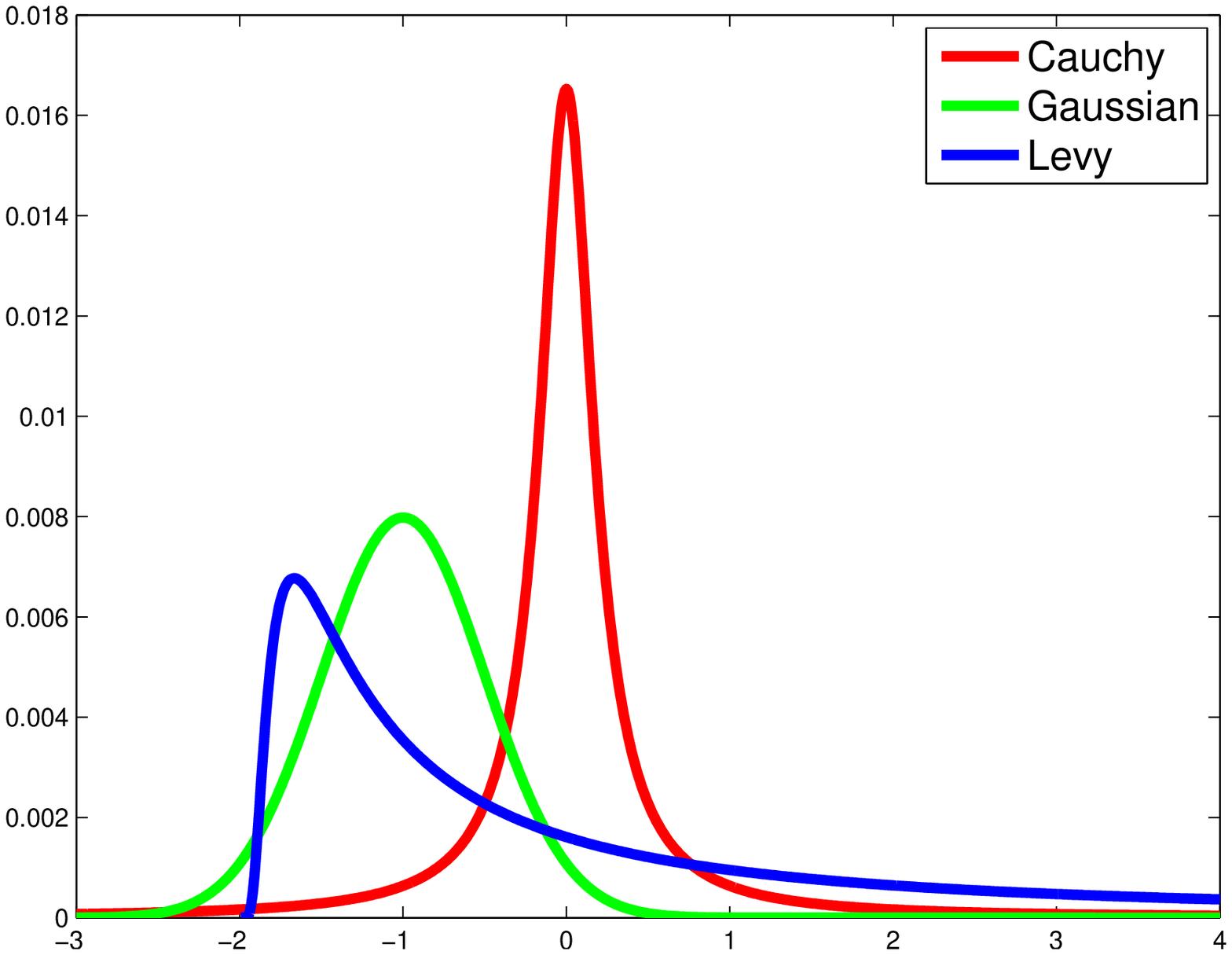}  
\end{minipage}
\label{stable-fig}
 \begin{picture}(0,0)
 \put(-380,-70){\small Algorithm 1:\ Exact inference in LCM using } 
\put(-380,-80){\small LCM-Elimination.} 
 \put(-180,-70){\small Figure 1:  The three special cases of stable}
  \put(-180,-80){\small distribution where closed-form pdf exists.}
 \end{picture} \\
\label{LCM-Elimination}
\vspace{1mm}
\end{figure}
\vspace{1mm}
\addtocounter{figure}{1}
Based on the results of Thm.~4.2 we propose an exact inference algorithm, LCM-Elimination, for computing the marginal cf (shown in Algorithm 1).\ We use the notation $N(k)$ as the set of graph neighbors of node $k$,  excluding $k$\footnote{More detailed explanation of the construction of a graphical model out of the linear relation matrix $A$ is found on \cite[Chapter 2.3]{phd-thesis}. }. $T$ is the set $\{t_1, \cdots, t_n\}$.

LCM-Elimination is dual to CFG-Elimination algorithm \cite{CFG}.  LCM-Elimination operates in the cf domain,  by evaluating one variable at a time, and updating the remaining graphical model accordingly.
The order of elimination does not affect correctness (although  it may affect efficiency). Once the marginal cf $\varphi(t_i)$, is computed, assuming the inverse Fourier transform exists, we can compute the desired marginal probability $p(x_i)$.

\subsection{Exact inference in stable distributions}
After defining LCM and showing that inference can be computed in the cf domain, we are finally ready to show how to compute exact inference in a linear model with underlying stable distributions. We assume that all observation nodes $Y_i$ are distributed according to a  stable distribution. 
From the linearity property  of stable distribution, it is clear that the hidden variables $X_i$ are distributed according to a stable distribution as well. 
The following theorem is one of the the novel contributions of this work, since as far as we know, no closed-form solution was previously derived.  
\begin{theorem}\label{sexact} Given a LCM, $Y=AX+Z$, with $n$ i.i.d. hidden variables $X_i \sim \stable{x_i},$  $n$ i.i.d. noise variables with known parameters $ Z_i\sim \stable{z_i},$ and $n$ observations $y_i\in \R$, assuming the matrix $A_{n\times n}$ is invertible\footnote{To simplify discussion we assume that the length of both the hidden and observation vectors $|X|=|Y|=n$. However the results can be equivalently extended to the more general case where $|X|=n, |Y|=m, m \ne n$. See for example \cite{ISIT2}.}, then\\ a) the observations $Y_i$ are distributed according to stable distribution $Y_i \sim \S(\alpha, \beta_{y_i}, \gamma_{y_i}, \delta_{y_i})$ with the following parameters: \small\[\gamma_{\vy}^\alpha =|\mA|^{\alpha}\gamma_{\vx}^\alpha+\gamma_{\vz}^\alpha ,
\ \ \  \beta_{\vy}=\gamma_\vy^{-\alpha}\odot[(|\mA|^\alpha\odot\sign(\mA))(\beta_\vx\odot \gamma_\vx)+\beta_{\vz}\odot \gamma_{\vz}] , \ \ \ \ \ \delta_{\vy} = \mA\delta_\vx+\xi_\vy\] \vspace{-2mm}
\BE \xi_\vy = \begin{cases} \tan(\tfrac{\pi \alpha}{2})[\beta_\vy\odot\gamma_{\vy}  -\mA(\beta_\vx  \odot \gamma_\vx)-\beta_\vz \odot \gamma_\vz] & \alpha \ne 1\\ 
\!\tfrac{2}{\pi}[\beta_{\vy}\odot\gamma_{\vy}\odot\log(\gamma_{\vy})-\mA^{}\odot\log(|\mA|)(\beta_\vx \odot \gamma_\vx)-\mA(\beta_\vx \odot \gamma_\vx\odot\log(\gamma_\vx))-\beta_\vz \odot \gamma_\vz] & \alpha = 1
\end{cases} \,,\vspace{-1mm}\nonumber \EE \normalsize 
 b) the result of exact inference for computing the marginals $ p(x_i|y)\ \sim \stable{x_i|y}$ is given in vector notation:
\small \BE {\beta_{x|y}} =\gamma_{x|y}^{-\alpha}\odot[(|A|^\alpha\odot \sign(A))^{-1}(\beta_y\odot\gamma_{y}^\alpha )]\,, \ \ \  {\gamma}_{x|y}^{\alpha} = (|A|^\alpha)^{-1}\gamma_{y}^\alpha,\ \ \ \ \  {\delta_{x|y}} =A^{-1}[\delta_{y} -\xi_x\,], \label{exactbeta} \EE\vspace{-2mm}
\BE  {\xi_{x|y}} = \begin{cases} \tan(\tfrac{\pi \alpha}{2})[\beta_y\odot\gamma_y-A(\beta_{{x|y}} \odot \gamma_{{x|y}})] & \alpha \ne 1\\ 
\!\tfrac{2}{\pi}[\beta_y\odot\gamma_y\odot\log(\gamma_y)-\!(A\odot\log(|A|)(\beta_{{x|y}} \odot \gamma_{{x|y}})-A(\beta _{{x|y}}\odot \gamma_{{x|y}}\odot \log(\gamma_{{x|y}}))] & \alpha = 1
\end{cases}\label{exactxi} \,,\vspace{-1mm}\EE
\normalsize
where  $\odot$ is the entrywise product (of both vectors and matrices),$|A|$ is the absolute value (entrywise)  $ \log(A), A^\alpha, \sign(A)$ are entrywise matrix operations and $\beta_x \triangleq [\beta_{x_1},\cdots,\beta_{x_n}]^T$ and the same for $\beta_y, \beta_z,\gamma_x, \gamma_y,\gamma_z, \delta_x, \delta_y, \delta_z.$
\end{theorem}

\vspace{-2mm}
\subsection{Approximate Inference in LCM}\vspace{-2mm}
Typically, the cost of exact inference may be expensive. For example, in the related linear model of a multivariate Gaussian (a special case of stable distribution), LCM-Elimination reduces to Gaussian elimination type algorithm with a cost of  $O(n^3),$ where $n$ is the number of variables. 
Approximate methods for inference like belief propagation \cite{BibDB:BookPearl}, usually require  less work than exact inference, but may not always converge (or convergence to an unsatisfactory solution). The cost of exact inference motivates us to devise a more efficient approximations.

We propose two novel algorithms that are variants of belief propagation for computing approximate inference in LCM. The first, Characteristic-Slice-Product (CSP)\ is defined in LCM (shown in Algorithm 2(a)). The second, Integral-Convolution (IC)\ algorithm (Algorithm 2(b)) is its dual in CFG. As in belief propagation, our algorithms are exact on tree graphical models.\ The following theorem establishes this fact.

\begin{theorem}\label{bp}
Given an LCM with underlying tree topology (the matrix $A$ is an irreducible adjacency matrix of a tree graph), the CSP and IC algorithms, compute exact inference, resulting in the marginal cf and the marginal distribution respectively. 
\end{theorem}

\begin{figure}[t!]
  \fbox{
 \begin{minipage}[b]{0.5\linewidth}
\scriptsize
{\tt Initialize:\ $m_{ij} (x_{j})= 1,\ \  \forall A_{ij} \ne 0$.\\Iterate until convergence}\\
\vspace{-2mm}
\[ m_{ij}(t_j) = \varphi_i(t_i,s_{1},\cdots,s_m) \!\!\!\prod_{k \in N(i)\setminus j}m_{ki}(t_i)\Bigg]_{t_i=0} \]
 {\tt Finally: \\
 \[ \varphi(t_{i})=\varphi_i(t_i,s_{1},\cdots,s_m)\prod_{k \in N(i)}m_{ki}(t_i). \]
}
 \vspace{-2mm}
 \end{minipage}
}
 \fbox{
 \begin{minipage}[b]{0.5\linewidth}
\scriptsize
{\tt Initialize:\ $m_{ij} (x_{j})= 1,\ \  \forall A_{ij} \ne 0$.\\
Iterate until convergence\\
\vspace{-2mm}
\[ m_{ij}(x_j) = \int \limits_{x_i} p(x_{i},y_1,\cdots,y_m)*\, \!\!\! \conv{k \in N(i)\setminus j}m_{ki}(x_i)dx_i\]
\vspace{-2mm}
 Finally: \\
 \[ p(x_{i})=p(x_{i},y_1,\cdots,y_m)*\conv{k \in N(i)}m_{ki}(x_i). \]
 \vspace{-2mm}
 }
 \end{minipage}
}
%

\fbox{
\begin{minipage}{5.7in}\small
\setlength{\baselineskip}{5mm} 
\scriptsize
{\tt Initialize:\ ${\beta}_{x_i|y}, {\gamma}_{x_i|y}, {\delta}_{x_i|y}= \S(\alpha, 0,0,0),\ \  \forall_{i}$.\\ 
\quad \quad Iterate until convergence:}\\ 
\vspace{-4mm}
\quad \begin{align*} {\gamma}_{x_i|y}^\alpha =\gamma_{y_i}^\alpha- \sum_{j \ne i}|A_{ij}|^\alpha\gamma_{x_j|y}^\alpha\,,  & &{\beta}_{x_i|y} = \beta_{y_{i}}\gamma_{y_i}^\alpha  - \sum_{j \ne i}\sign(A_{ij})|A_{ij}|^\alpha\beta_{x_j|y}\,, & & \delta_{x_i|y}=\delta_{y_i}-\sum_{j \ne i}A_{ij}\delta_{x_j|y}\,-\xi_{x_i|y}, \end{align*}\vspace{-3mm}
\BE \hspace{-1cm}{\xi}_{x_i|y} = \begin{cases}\tan(\tfrac{\pi \alpha}{2})[\beta_{y_i}\gamma_{y_i}-\sum_{j}A_{ij}\beta_{x_j|y}\gamma_{x_j|y}^{\tfrac{1-\alpha}{\alpha}}]& \alpha \ne 1\\
\tfrac{2}{\pi}[\beta_{y_i}\gamma_{y_i}\log(\gamma_{y_i})-\sum_{j:A_{ij}\ne0  }A_{ij}\log(|A_{ij}|)\beta_{x_j|y}\gamma_{x_j|y}^{\tfrac{1-\alpha}{\alpha}}-\sum_{j  }A_{ij}\beta_{x_j|y}\gamma_{x_j|y}\log(\gamma_{x_j|y}^{\tfrac{1-\alpha}{\alpha}})] & \alpha = 1 \end{cases} \label{jacobi3}\EE
{\tt Output: $x_i|y\sim \S(\alpha, {\beta}_{x_i|y}/{\gamma_{x_i|y}^\alpha}, {\gamma}_{x_i|y}, {\delta}_{x_i|y})$}
\end{minipage} 
}
 \begin{picture}(0,0)
 \put(-420,50){(a)}
 \put(-210,50){(b)}
 \put(-210,-36){(c)}
 \end{picture}\vspace{2mm} \\
\small Algorithm 2: Approximate inference in LCM using the (a)\ Characteristic-Sum-Product (CSP)\ algorithm (b)\ Integral Convolution (IC) algorithm. Both are exact on tree topologies. (c) Stable-Jacobi algorithm.\ 
\label{algos}\vspace{-2mm}\vspace{-2mm}\vspace{-2mm}
\end{figure}\vspace{-2mm}
The basic property which allows us to devise the  CSP algorithm is that LCM is defined as a product of factor in the cf domain. Typically, belief propagation algorithms are applied to a probability distribution which factors as a product of potentials in the pdf domain. The sum-product algorithm uses the distributivity of the integral and product operation to devise efficient recursive evaluation of the marginal probability. Equivalently, the Characteristic-Slice-Product algorithm uses the distributivity of the slicing and product operations to perform efficient inference to compute the marginal cf in the cf domain, as shown in Theorem 4.4. In a similar way, the Integral-Convolution algorithm uses distributivity of the integral and convolution operations to perform efficient inference in the pdf domain. Note that the original CFG work \cite{CFG0,CFG} did not consider approximate inference. Hence our proposed approximate inference algorithm further extends the CFG model.     
\vspace{-2mm}
\subsection{Approximate inference for stable distributions}\label{approx_inf}\vspace{-2mm}
For the  case of stable distributions, we derive an approximation algorithm, Stable-Jacobi (Algorithm~2(c)), out of the CSP update rules.  The algorithm is derived by substituting the convolution and multiplication by scalar operations (Prop.~\ref{prop_s2} b,a) into the update rules of the CSP\ algorithm given in Algorithm~2(a).

Like belief propagation, our approximate algorithm Stable-Jacobi is not guaranteed to converge on general graphs containing cycles.
We have analyzed the evolution dynamics of the update equations for Stable-Jacobi and derived sufficient conditions for convergence. Furthermore, we have analyzed the accuracy of the approximation. Not surprisingly, the sufficient condition for convergence  relates to the properties of the linear transformation matrix $A$. The following theorem is one of the main novel contributions of this work. It provides both sufficient condition for convergence of Stable-Jacobi as well as closed-form equations for the fixed point. 



\begin{theorem}\label{sconv}
Given a LCM with $n$  i.i.d  hidden variables $X_i$,  $n$ observations $Y_i$ distributed according to stable distribution $Y_i \sim \S(\alpha, \beta_{y_i}, \gamma_{y_i}, \delta_{y_i}),$ assuming the linear  relation matrix $\mA_{n\times n}$ is invertible and normalized to a unit diagonal\footnote{When the matrix $\mA$ is positive definite it is always possible to normalize it to a  unit diagonal. The normalized matrix is $\mD^{-\tfrac{1}{2}}\mA\mD^{-\tfrac{1}{2}}$ where  $\mD = \diag(\mA)$. Normalizing to a unit diagonal is done to simplify convergence analysis (as done for example in \cite{WS}) but does not limit the generality of the proposed method.}, Stable-Jacobi (as given in Algorithm 2(c))\ converges to a unique fixed point under both the following sufficient conditions for convergence (both should hold): 
\begin{align*}\small
(1)& &\rho(|R|^\alpha)<1\,, & & &(2)& \rho(R) < 1\,.
\end{align*}
where $\rho(R)$ is the spectral radius (the largest absolute value of the eigenvalues of $R$),  $R \triangleq I-A,$  $|R|$ is the entrywise absolute value and $|R|^\alpha$ is the entrywise exponentiation. Furthermore, the unique fixed points of convergence are given by equations \eqref{exactbeta}-\eqref{exactxi}. The algorithm converges to the {\bf exact marginals} for the linear-stable channel.\footnote{Note that there is an interesting relation to the walk-summability convergence condition \cite{WS} of belief propagation in the Gaussian case:  $\rho(|R|)<1$. However, our results are more general since they apply for any characteristic exponent $0<\alpha\le2$ and not just for $\alpha=2$ as in the Gaussian case.}
\end{theorem}

\vspace{-2mm} 
\section{Application:\ Network flow monitoring}\vspace{-2mm}
In this section we propose a novel application for inference in LCMs to model network traffic flows of a large operational worldwide testbed. Additional experimental results using synthetic examples are found in the supplementary material. Network monitoring is an important problem in monitoring and anomaly detection of communication networks \cite{Flows1,Flows3,ATANDT}. We  obtained Netflow PlanetLab network data \cite{NETFLOW} collected on 25 January 2010. The PlanetLab network \cite{PLANETLAB} is a distributed networking testbed with around 1000 server nodes scattered in about 500 sites around the world. 
We define a network flow as a directed edge between a transmitting and receiving hosts. The number of packets transmitted in this flow is the scalar edge weight.  

We propose to use LCMs for  modeling distribution of network flows. Figure 2(a) plots a distribution of flows, sorted by their bandwidth, on a typical PlanetLab node. Empirically, we found out that network flow distribution in a single PlanetLab node are fitted quite well using L\'evy distribution a stable distribution with $\alpha=0.5, \beta=1.$ The empirical means are  $\mathtt{mean}(\gamma)\approx 1e^{-4} $, $\mathtt {mean}(\delta)\approx 1$. For performing the fitting, we use Mark Veillette's Matlab stable distribution package \cite{MV}.

\begin{figure}
\begin{minipage}[b]{0.33\linewidth}
\begin{center}
\includegraphics[scale=0.28,clip]{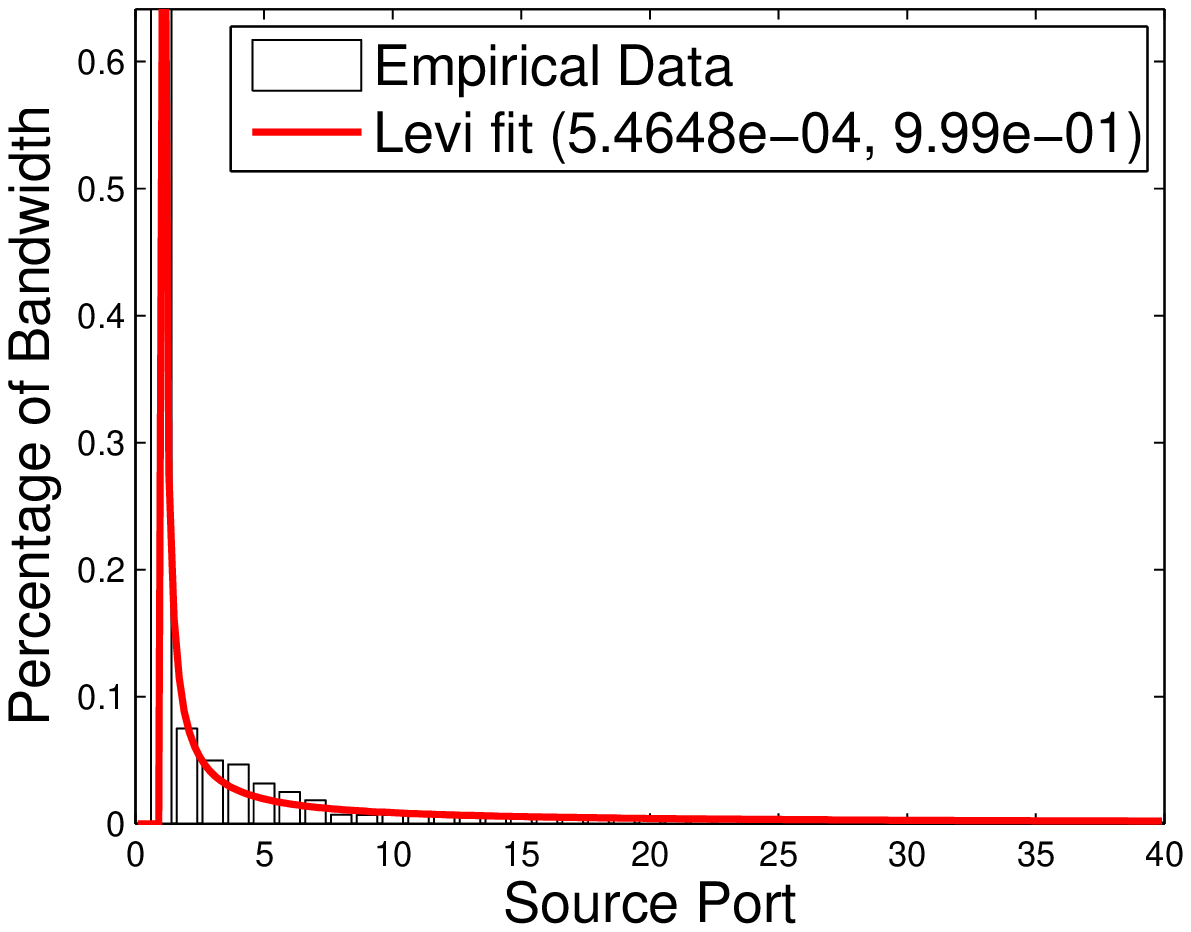}
(a)
\hspace{3mm}
\end{center}
\end{minipage}
\begin{minipage}[b]{0.33\linewidth}
\centering{
\includegraphics[clip,scale=0.23,bb=61 108 539 476]{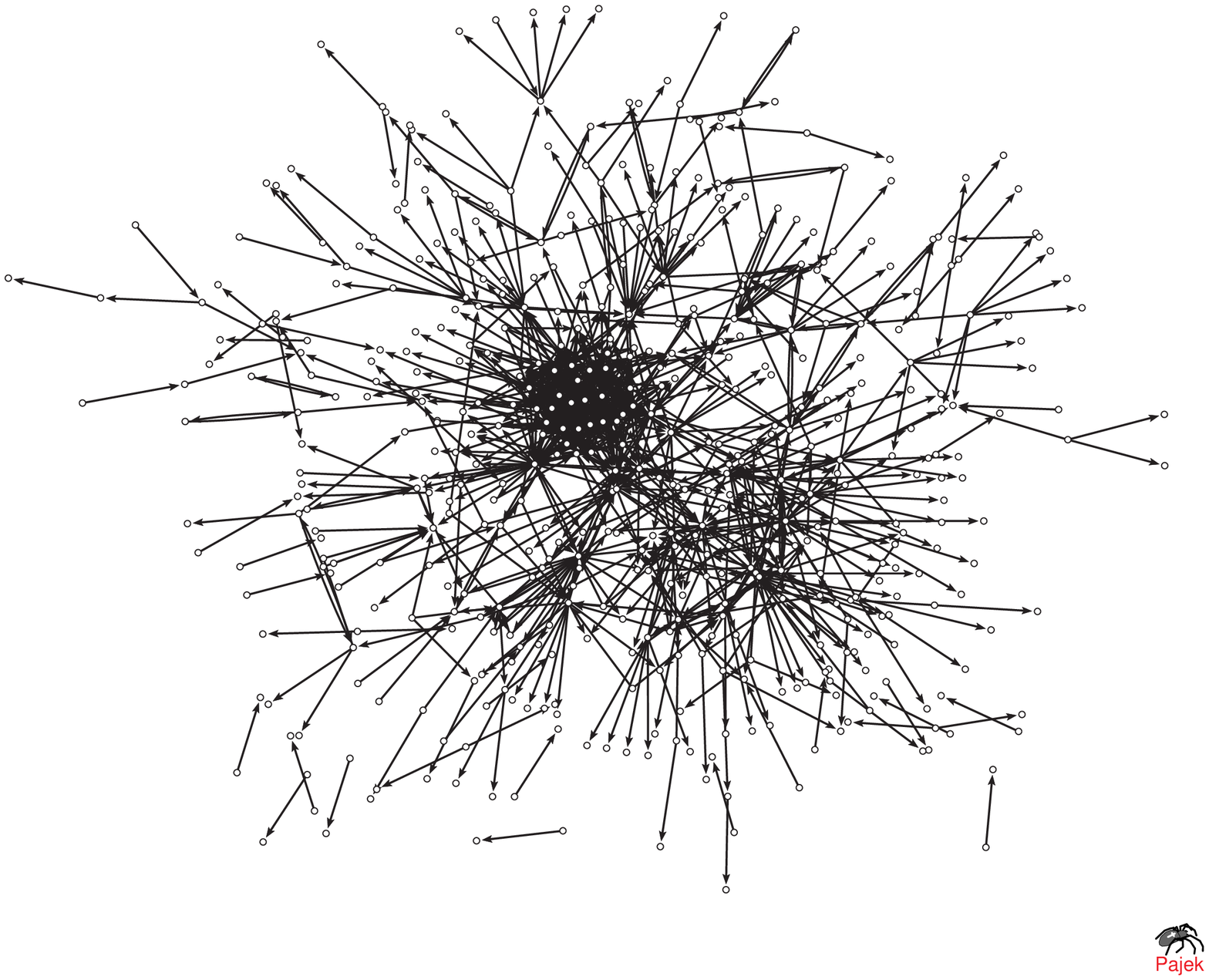}
(b)
}
\end{minipage}
\begin{minipage}[b]{0.33\linewidth}
\centering{
\includegraphics[clip,scale=0.28]{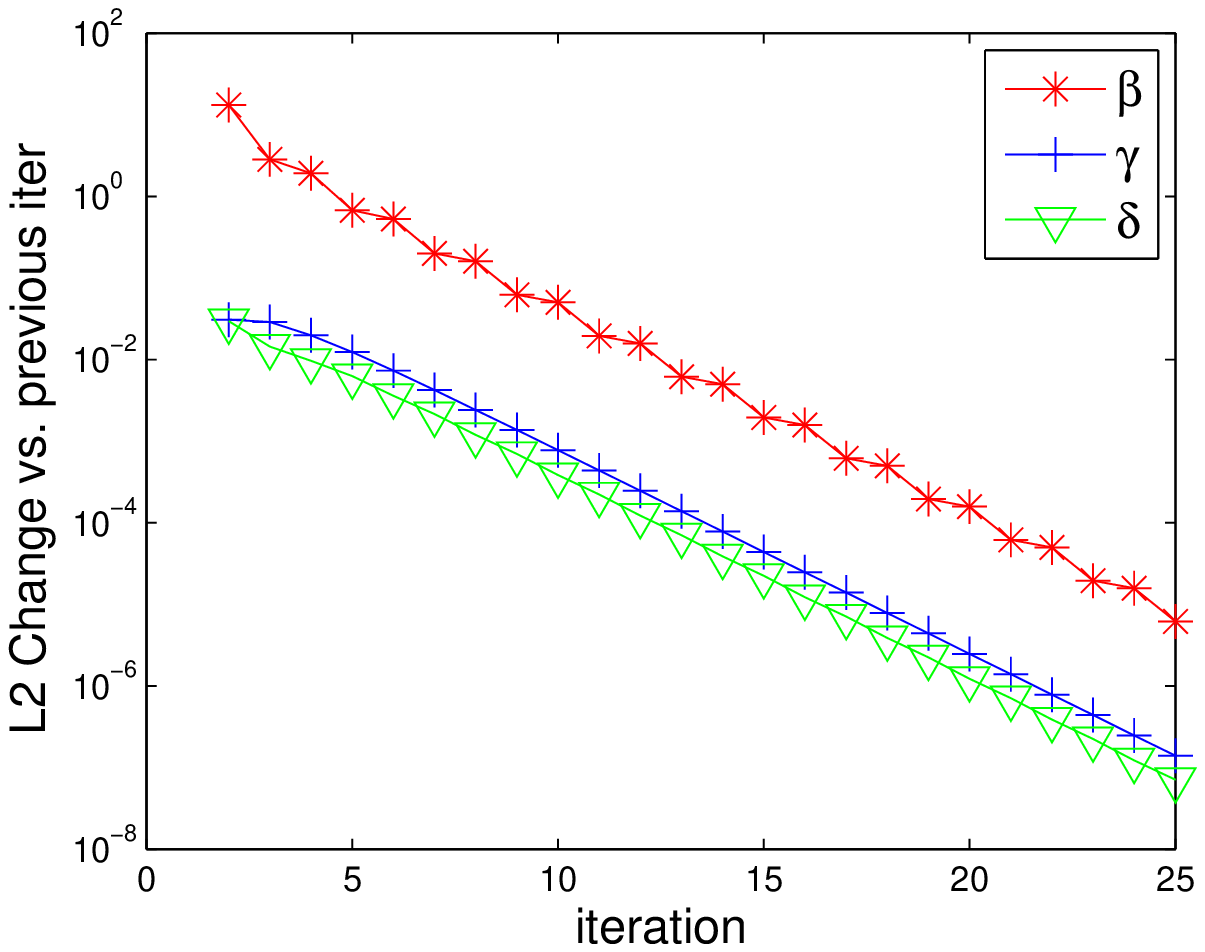}
(c)
}
\end{minipage}
\label{fig:exp}

\caption{\small(a) Distribution of network flows on a typical PlanetLab host is fitted quite well with a Levy distribution. (b) The core of the PlanetLab network. 1\% of the flows consists of 19\% of the total bandwidth. (c) Convergence of Stable-Jacobi. \ }
\vspace{-6mm}
\end{figure}
 Using previously proposed techniques utilizing histograms \cite{Flows3} for tracking flow distribution in Figure 2(a), we would need to store 40 values (percentage of bandwidth for each source port). In contrast,  by approximating network flow distribution with stable distributions, we need only 4 parameters ($\alpha,\beta,\gamma,\delta)$! Thus we dramatically reduce storage requirements. Furthermore, using the developed theory in previous sections, we are able to linearly aggregate distribution of flows in clusters of nodes.

We extracted a connected component of traffic flows connecting the core network $652$ nodes. We fitted a stable distribution characterizing flow behavior for each machine. A partition of $376$ machines as the observed flows $Y_i$ (where flow distribution is known). The task is to predict the distribution of the unobserved remaining $376$ flows $X_i$, based on the observed traffic  flows (entries of $A_{ij}$). We run approximate inference using Stable-Jacobi and compared the results to the exact result computed by LCM-Elimination. We emphasize again, that using related techniques
 (Copula method ,  CFG, and ICA)\ it is not possible to compute exact inference for the problem at hand. 
In the supplementary material, we provide a detailed comparison of two previous  approximation algorithms: non-parametric BP (NBP) and expectation propagation (EP). 

Figure 2(c) plots convergence of the three parameters $\beta, \gamma, \delta$ as a function of iteration number of the Stable-Jacobi algorithm. Note that convergence speed is geometric. ($\rho(R)=0.02 << 1$).
Regarding computation overhead, LCM-Exact algorithm requires $4\cdot376^3$ operations, while Stable-Jacobi converged to an accuracy of $1e^{-5}$ in only $4\cdot376^2 \cdot 25$ operations. Additional benefit of the Stable-Jacobi  is that it is a distributed algorithm, naturally suitable for communication networks.
Source code of some of the algorithms presented here can be found on \cite{STABLECODE}.

%
%

%
\vspace{-3mm}
\section{Conclusion and future work}\vspace{-2mm}
We have presented a  novel linear graphical model called LCM, defined in the cf domain. We have shown for the first time how to perform exact and approximate inference in a linear multivariate graphical model when the underlying distributions are stable. We have discussed an application of our construction for computing inference of network flows. 

We have proposed to borrow ideas from belief propagation, for computing efficient inference, based on the distributivity property of the slice-product operations and the integral-convolution operations. We believe that other problem domains may benefit from this construction, and plan to pursue this as a future work. 

We believe there are several exciting directions for extending this work. Other families of distributions like geometric stable distributions or Wishart  can be analyzed in our model. The Fourier transform can be replaced with more general kernel transform, creating richer models.   
\vspace{-4mm}
\section*{Acknowledgement}
\vspace{-2mm}
\small
D. Bickson would like to thank Andrea Pagnani (ISI)\ for inspiring the direction of this research, to John P. Nolan (American University)\ , for sharing parts of his excellent book about stable distribution online, Mark Veillette (Boston University)\ for sharing his stable distribution code online, to Jason K. Johnson (LANL) for assisting in the convergence analysis and to Sapan Bathia and Marc E. Fiuczynski (Princeton University)\ for providing the PlanetFlow data. This research was supported by ARO MURI W911NF0710287, ARO MURI W911NF0810242, NSF Mundo IIS-0803333 and NSF Nets-NBD  CNS-0721591.
\footnotesize
\bibliographystyle{abbrv}
\vspace{-2mm}\bibliography{references}\vspace{-3mm}
\onecolumn
\section{Supplementary material}
\normalsize

\begin{defs} {\bf Characteristic Function.} For a scalar random variable $X$, the {\em characteristic function} is defined as the expected value of $e^{i t X}$ where $i$ is the imaginary unit, and $t \in \R$ is the argument of the characteristic function: $\varphi_X(t) = E[e^{i t X}] = \int_{- \infty}^{\infty}e^{i t x}dF_X(x)$
where $F_X(x)$ is the cumulative distribution function of $X$. If a random variable $X$ has a probability density function $f_X$, then the characteristic function is its Fourier transform, $\varphi_X(t) = \int \limits_{- \infty}^{\infty} e^{itx}f_X(x) dx$.
\end{defs}

\subsection{Proof of Theorem \ref{duality}}
\begin{proof}\small
\[ \F(p(x,y)) = \F( \conv{i} (p(x_{i},y_1,\cdots,y_m))= \prod_i \F(p(x_i,y_1,\cdots,y_m))= \] \[ =\prod_i \varphi(t_{i},s_{1},\cdots,s_m)=  \varphi(t_1,\cdots,t_n,s_1,\cdots,s_m). \]
\end{proof}
\subsection{Proof of Theorem \ref{duality2}}
\begin{proof}
\BES\ \F^{-1}(\varphi(t_1,\cdots,t_n,s_1,\cdots,s_m))=\F^{-1}(\prod_i \varphi(t_{i},s_{1},\cdots,s_m)) \EES
\BES =\conv{i}\F^{-1} (\varphi(t_{i},s_{1},\cdots,s_m))\ = \conv{i} p(x_{i},y_1,\cdots,y_m)  =p(x,y)\,.\EES
\end{proof}

\subsection{Proof of Theorem \ref{lcm3}}
\begin{proof}
The proof follows from the Projection-Slice theorem (also known as the Central Slice theorem)\ \cite[p. 349]{SliceProjection}, which is briefly stated here. Let $f(x,y)$ be a multivariate function and $F(u,v )$ be its matching Fourier transform. Then 
\small
\[ \F\{f(x)\}=\F\{\int_{-\infty}^\infty f(x,y)dy\}=\int_{-\infty}^\infty e^{iux}[ \int_{-\infty}^\infty f(x,y)dy]dx= \int_{-\infty}^\infty  \int_{-\infty}^\infty e^{iux}f(x,y)dx dy=F(u,0)\,.\] \normalsize
This theorem is naturally extended to multiple variables. In our case, 
\BES \varphi(0,0, t_{i,}\cdots,0)= \prod_j \varphi(t_j,s_{1},\cdots,s_m) \Bigg]_{T \setminus i=0} =  \int_{-\infty}^\infty e^{it_ix_{i}} \big[  \conv{j}p(x_j,y_1, \cdots,y_m)\ \big] d_{X} = \EES
\BES  \int_{-\infty}^\infty e^{it_ix_{i}}  \big[ \int_{-\infty}^\infty  \conv{j}p(x_j,y_1, \cdots,y_m)d_{X\setminus i}  \ \big] dx_i=  \F\{ \int \limits_{X\setminus i}\big[\conv{j}p(x_j,y_1, \cdots,y_m)\big] d_{X\setminus i}\}=\F\{p(x_i)\}\,. \EES
\normalsize 
\end{proof}

\subsection{Proof of Theorem \ref{sexact}}
\normalsize
\begin{proof}
For simplicity, we do not handle the noise variable $z$ in this proof. The noise can be added as a regularization later. We use the linear  relation between distributions to extract $X$: $X=A^{-1}Y$. 
Note that $X$ must distribute according to stable distribution since it is composed from linear combination of stable variables. For the scale parameter we get
(using the linearity of $A$ substituted in Prop. \ref{prop_s2} (a),(b))
\BE\gamma_{y_i}^\alpha = \sum_{j}|\mA_{ij}|^\alpha\gamma_{x_j}^\alpha \nonumber \EE
In vector notation we got
\[ \gamma_y^\alpha=|\mA|^\alpha\gamma_x^\alpha \,.\]
Solving this  linear system of equations we get \[\gamma_{x|y}^\alpha=(|\mA|^\alpha)^{-1}[\gamma_y^\alpha].\]
Regarding the skew parameter $\beta_\vx$ using Prop. \ref{prop_s2}(a,b) we get that 
\[
\beta_{y_i} = \frac{\sum_j\sign(A_{ij})|A_{ij}|^\alpha\beta_{x_j|y}\gamma_{x_j|y}^\alpha}{\gamma_{y_i}^\alpha}\,.\]
In vector notation we get
\[ \beta_y =\gamma_y^{-\alpha}\odot[(\sign(\mA)\odot|\mA|^\alpha)(\beta_{x}\odot\gamma^\alpha_x)]\,. \]
Now assume that $\gamma_x^\alpha$ is a known constant, we can exact $\beta_x $ and get
\[ \beta_{\vx|y} =\gamma^{-\alpha}_{\vx|\vy}\odot[((\sign(\mA)\odot|\mA|^\alpha)^{-1}(\beta_\vy\odot\gamma^\alpha)]\,. \nonumber \]
Regarding the location parameter $\delta_\vx$,
\[ \delta_{y_i} = \sum_j A_{ij}\delta_{x_j}+\xi_{i}\,, \]
\[ \xi_i=
\begin{cases} \tan(\tfrac{\pi \alpha}{2})[\beta_{y_{i}}\gamma_{y_{i}}-\sum_j\sign(A_{ij})|A_{ij}|\beta_{x_j}\gamma_{x_j}]& \alpha \ne 1\\
\tfrac{2}{\pi}[\beta_{y_{i}}\gamma_{y_{i}}\log(\gamma_{y_{i}})-\sum_j\sign(A_{ij})|A_{ij}|\beta_{x_j}\gamma_{x_j}\log(|A_{ij}|\gamma_{x_j})] & \alpha = 1\end{cases}\,. \]
In matrix notation (after some algebra)\ we get
\[ \delta_{\vy} = \mA\delta_\vx +\xi \]
\[ \xi =
\begin{cases} \tan(\tfrac{\pi \alpha}{2})[\beta_{\vy}\odot\gamma_{\vy}-\mA ( \beta_{\vx}\odot\gamma_{\vx})]& \alpha \ne 1\\
\tfrac{2}{\pi}[\beta_{\vy}\odot\gamma_{\vy}\odot\log(\gamma_{\vy})-(\mA\odot\log(|\mA|)) ( \beta_{\vx}\odot\gamma_{\vx})-\mA ( \beta_{\vx}\odot\gamma_{\vx}\odot\log(\gamma_{\vy}))] & \alpha = 1\end{cases}\,. \]
In total we got a linear system that is solved using
\[ \delta_{\vx|y} = \mA^{-1}(\delta_\vy-\xi)\,.\]
\end{proof}

\subsection{Proof of Theorem \ref{bp} }

\begin{proof}W.l.g we prove for the Slice-product algorithm 2(a). The other algorithms are symmetric because the slice/convolution and integral/convolution operations maintain the distributivity property as well.   

We are interested in computing the posterior marginal probability\BE p(x_i)=\int \limits_{\vx\setminus i}p(\vx,\vy)d_{X \setminus i} \sim\ \int \limits_{X\setminus i}p(x_1, \cdots, x_n,y_{1},\cdots,y_m)d_{X \setminus i}\EE \BE= \F^{-1}\{\prod_{i}\varphi(t_{i},s_1,\cdots,s_m)\Big]_{t_i=0}\}\,.  \label{correct} \EE
 W.l.g assume that $X_i$ is a tree root.\ Its matching marginal cf $\varphi(t_i)$ can be written as a combination of incoming message computed by the neighboring sub trees:
\BE \varphi(t_i) \sim \varphi(t_{i},s_1,\cdots,s_m) \prod_{j\in N(i)}m_{ji}(t_i)\,, \nonumber \EE
where the messages $m_{ji}(t_i)$ are defined by the algorithm 2(a).
We prove using full induction on the tree diameter. The messages $m_{ji}(t_i)$ satisfy the recursion:
\[ m_{ji}(t_i) = \varphi(t_{i},s_1,\cdots,s_m)\, \!\!\! \prod_{k \in N(j)\setminus i}m_{kj}(x_j)\Big]_{x_{j=0}}.\]The basis for the induction is a tree with a single node $x_{1}$. In this case there are no incoming messages,  $\varphi(t_1)=\varphi(t_{i},s_1,\cdots,s_m)$ and we are done. Now assume that the induction assumption holds for a tree with diameter $d-1$ or less and we want to prove it for a tree with diameter $d$.
We make the following construction. We add a new node $x_i$ to the tree to get a tree with diameter $d$. This node has one or more neighbors $j\in N(i)$. \[ \varphi(t_i) \sim \varphi(t_{i},s_1,\cdots,s_m) \prod_{j\in N(i)}m_{ji}(t_i)\,= \]
\[=(t_{i},s_1,\cdots,s_m) \prod_{j\in N(i)} p(t_{j},s_1,\cdots,s_m)\, \!\!\! \prod_{l \in N(j)\setminus i}m_{lj}(t_j)]\Big]_{t_j=0}\,. \]
Using distributivity of the slice/product (algorithm 2(a)), and the tree assumption (separate trees connected to  node $k$ are disjoint), we interchange order of operators to get:
\[ \varphi(t_i) \sim\varphi(t_{i},s_1,\cdots,s_m)[ \prod_{j\ne i}\varphi(t_{j},s_1,\cdots,s_m))]\Big]_{t_{j\ne i}=0}=\, \!\!\! \]
\[ = \prod_{i}\varphi(t_{i},s_1,\cdots,s_m)\Big]_{t_{X\setminus i}=0}\]
This completes the proof since we have obtained the formulation \eqref{correct}.
\end{proof}




\label{gen_inst}

\ignore{
\subsection{Derivation of LCM-Approx}
Following we provide the basic derivation of the Stable-Approx algorithm (shown in Algorithm 2(c)). 

Assume the stable exponent $\alpha$ is fixed and common to  all nodes.
We initialize the initial message to be  $m_{ij} = \S(\alpha,0,0,0)$. We use the notation $m_{ij}(t_j) = \S(\alpha, \beta_{ij}, \gamma_{ij}, \delta_{ij})$ to present the three parameters $( \beta_{ij}, \gamma_{ij}, \delta_{ij})$ passed on the edge $(i,j)$. We start with the update rule:
 \BE m_{ij}(t_j) =\varphi(t_i,s_{1},\cdots,s_n) \!\!\!\prod_{k \in N(i)\setminus j}m_{ki}(t_i)]\Bigg]_{t_i=0} \label{int_stab}\EE

We evaluate the product step  $\varphi(t_i,s_{1},\cdots,s_n) \prod_{k \in N(i)\setminus j}m_{ki}(t_i)$
using the multivariate version of Prop. \ref{prop_s2} b:
\small
\BE \gamma_{i \setminus j}^\alpha = \gamma_i^\alpha+\sum_{k \in N(i) \setminus j} \gamma_{ki}^\alpha\label{cs1}\,, \EE
\BE \beta_{i \setminus j} ~ = \gamma_{i \setminus j}^{-\alpha} (\beta_i\gamma_i^\alpha+\sum_{k \in N(i) \setminus j}\beta_{ki}\gamma_{ki}^\alpha)\label{cs2}\,, \EE
\BE \delta_{i \setminus j} =\delta_i+\sum_{k \in N(i) \setminus j} \delta_{ki}+\xi_{k\setminus i}\,, \label{cs3} \EE
\BE \xi_{i \setminus j}=\begin{cases} \tan(\tfrac{\pi \alpha}{2})[\beta_{i \setminus j}\gamma_{i \setminus j}-\beta_i \gamma_i-\sum_{k \in N(i) \setminus j}\beta_{ki}\gamma_{ki}]& \alpha \ne 1\\
\tfrac{2}{\pi}[\beta_{i \setminus j}\gamma_{i \setminus j}\log\gamma_{i \setminus j}-\beta_i \gamma_i -\sum_{k \in N(i) \setminus j}\beta_{ki}\gamma_{ki}\log(\gamma_{ki})] & \alpha = 1\end{cases}\\ \label{cs4}\,. \EE
\normalsize
Now we perform slicing/evaluation using Prop. \ref{prop_s2} (a)\ and get

\BE \gamma_{ij} =  -|A_{ij}|\gamma_{i \setminus j}\,,\label{cs5}\EE
\BE \beta_{ij} ~ = -\sign(A_{ij}) \beta_{i \setminus j}\label{cs6}\,,\EE
\BE \delta_{ij} =-A_{ij}\delta_{i \setminus j}\,. \label{cs7}\EE

In total we have defined update rules for computing the CSP-Stable message $m_{ij}(t_j) = \S(\alpha, \beta_{ij}, \gamma_{ij}, \delta_{ij}).$
The algorithm as defined by equations \eqref{cs1}-\eqref{cs7}
is guaranteed to converge on tree topologies, to the exact result (as shown in Theorem \ref{bp}). When the underlying topology is loopy, it is not clear whether it will converge and what will be the accuracy of the result. Unfortunately its convergence analysis is rather complex. 

We have simplified a little the model, creating a  variant named Stable-Approx which is presented in Algorithm 2(c). Stable-Approx is derived by substituting \eqref{cs1}-\eqref{cs4} into \eqref{cs5}-\eqref{cs7}. The only difference between the two algorithms is that on the later, summation \eqref{cs1}-\eqref{cs4} is performed using all the neighbors $k \in N(i)$ (not excluding neighbor $j$). In the Following, we give sufficient conditions for the convergence of Stable-Approx. For simplicity of exposition, we assume that the graphical model is normalized s.t. $\gamma_i = 1, \forall_i$.
} 

 \subsection{Proof of Theorem \ref{sconv}}
 \begin{proof}
We start with the scale parameter calculation since it is decoupled from the other parameters. 
\[ {\gamma}_{x_i|y}^\alpha =\gamma_{y_i}- \sum_{j \ne i}\gamma_{x_j|y}^\alpha |A_{ij}|^\alpha \]
This iteration is a Jacobi iteration for solving the linear system
\[ |\mA|^\alpha\gamma_\vx^\alpha =\gamma_{\vy}^\alpha \]
The linear system solution is given in \eqref{exactbeta} as desired.
It is further known that this iteration converges when $\rho(|R|^\alpha)<1$.

Regarding the skew parameter $\beta$ the Stable-Jacobi update rule is:
\[ \beta_{x_i|y} =\beta_{y_i} \gamma_{y_i}^\alpha- \sum_{j \ne i}\sign(A_{ij})|A_{ij}|^\alpha \beta_{x_j|y}\,. \]
This conforms to the Jacobi equation for solving the linear system
\[ [|\mA|^\alpha\odot \sign(\mA)]\beta_\vx = \beta_\vy\odot\gamma_\vy^\alpha  \]
Assuming this system converged, we divide by $\gamma_\vx^\alpha $ to get \eqref{exactbeta}
\[ \beta_{\vx|\vy} =\gamma_{\vx|\vy}^{-\alpha}\odot[|\mA|^\alpha\odot \sign(\mA)]^{-1}[\beta_\vy\odot\gamma_\vy^\alpha ]\,. \]
The iteration for computing a skew parameter $\beta$ converges when $\rho(|\mR|^\alpha\odot \sign(\mR))<1$. Using \cite[Theorem 8.4.5, Section 8.4]{MatrixAnalysis}
we get that 
$\rho(|\mR^\alpha\odot \sign(\mR)|) = \rho(|\mR^\alpha|) >  \rho(|\mR|^\alpha\odot \sign(\mR))$. In other words, when the sufficient condition for the scale parameter $\gamma$ holds ($\rho(|\mR^\alpha|)<1$), then the skew parameter $\beta$  converges as well. 

Now we analyze the shift parameter $\delta$ evolution. The parameter is given by 
\[ {\delta}_{x_i|y} =\delta_{y_{i}}-\sum_{j \ne i}A_{ij}\delta_{x_j|y}-\xi_{x_j|y}\,, \]
This is a Jacobi equation for solving the linear system 
\[ \mA\delta_\vx  = \delta_\vy- \xi_x \ \]
Is given in \eqref{exactbeta}.
This iteration converges when $\rho(\mR)<1$, which is the second sufficient condition for convergence.
\end{proof}

\subsection{Synthetic example}
\begin{figure*}[h!] 
  \begin{center}
    \subfigure[Convergence of message norms]{
     \includegraphics[width=.3\textwidth]{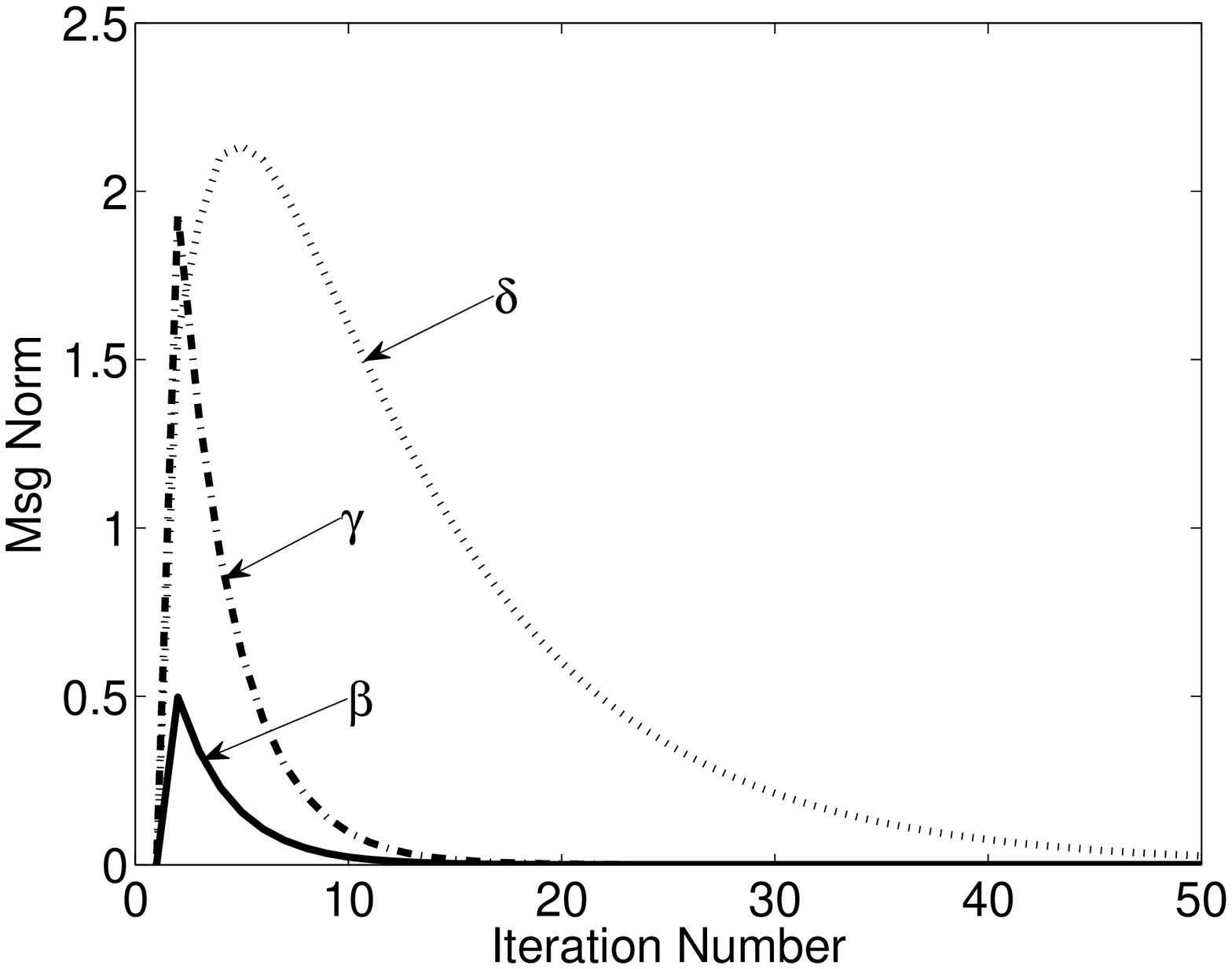}
    }
     \subfigure[Distance to true solution]{
      \includegraphics[width=.3\textwidth]{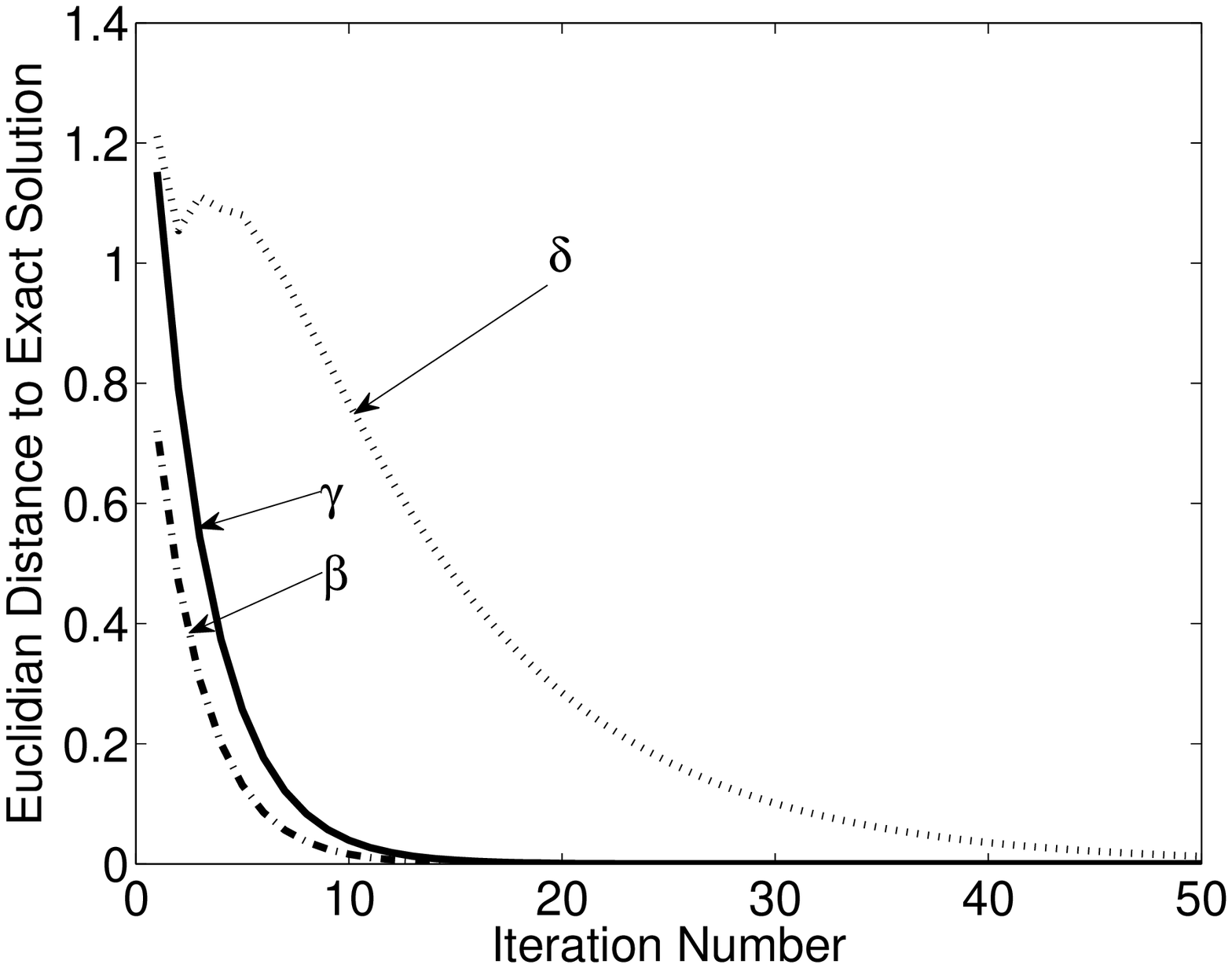}
    }
     \subfigure[Distance to true solution (log scale)]{
      \includegraphics[width=.3\textwidth]{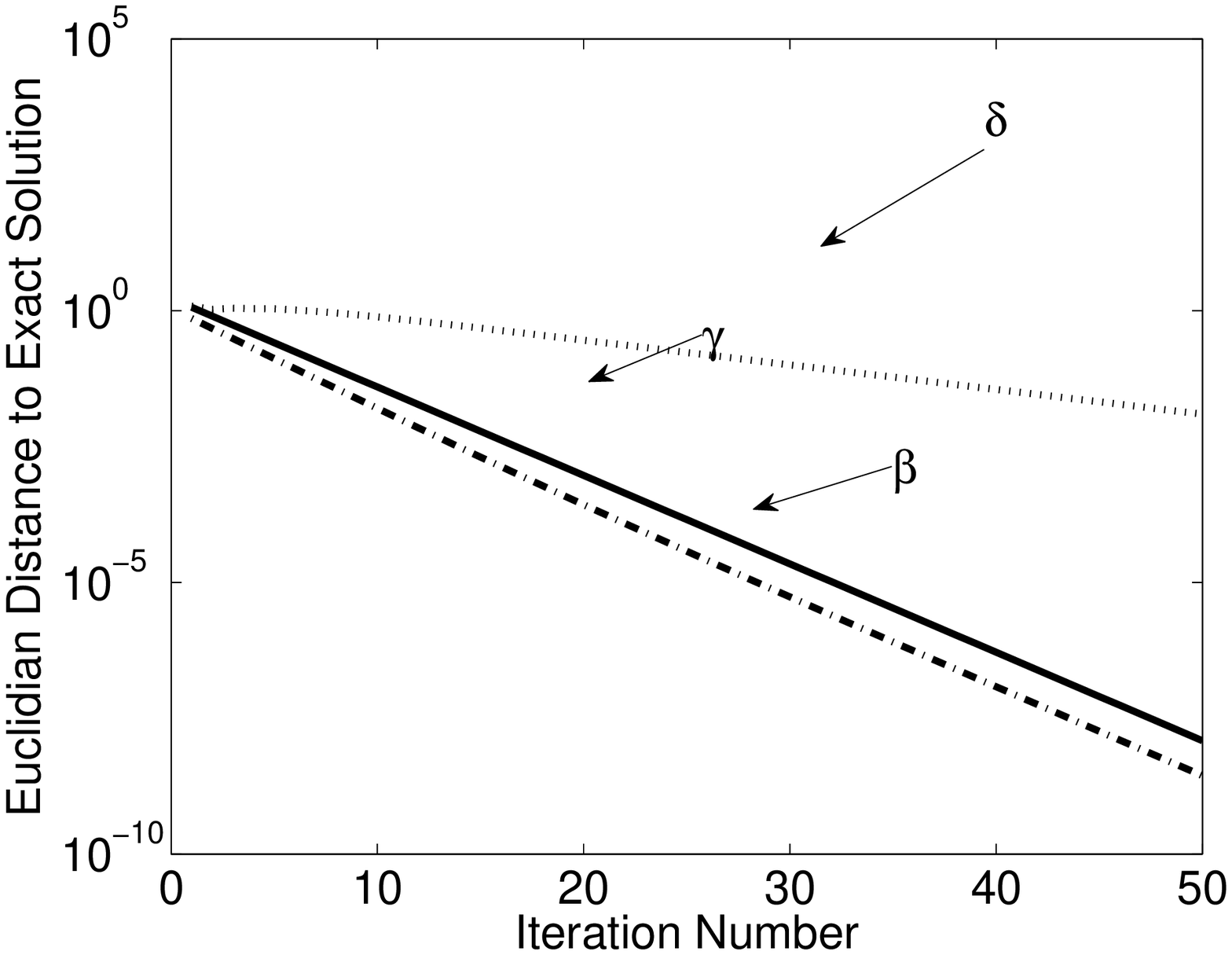}
     }
    \subfigure[Posterior marginal of $x_2$]{
      \label{fig:CoemScalabilityPlotb}
      \includegraphics[width=.3\textwidth]{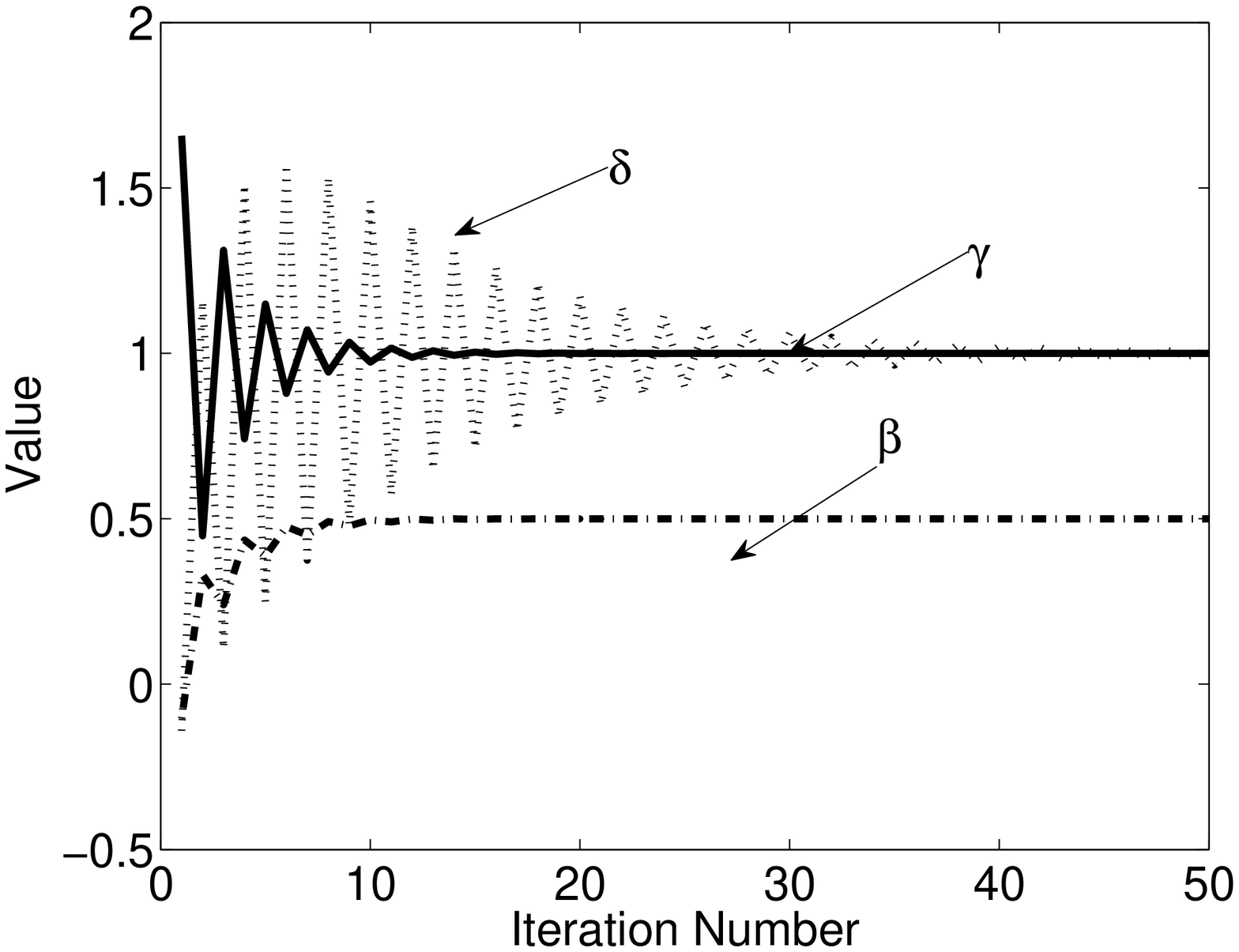}
    }
    \subfigure[Distance to true solution of $x_2$]{
      \includegraphics[width=.3\textwidth]{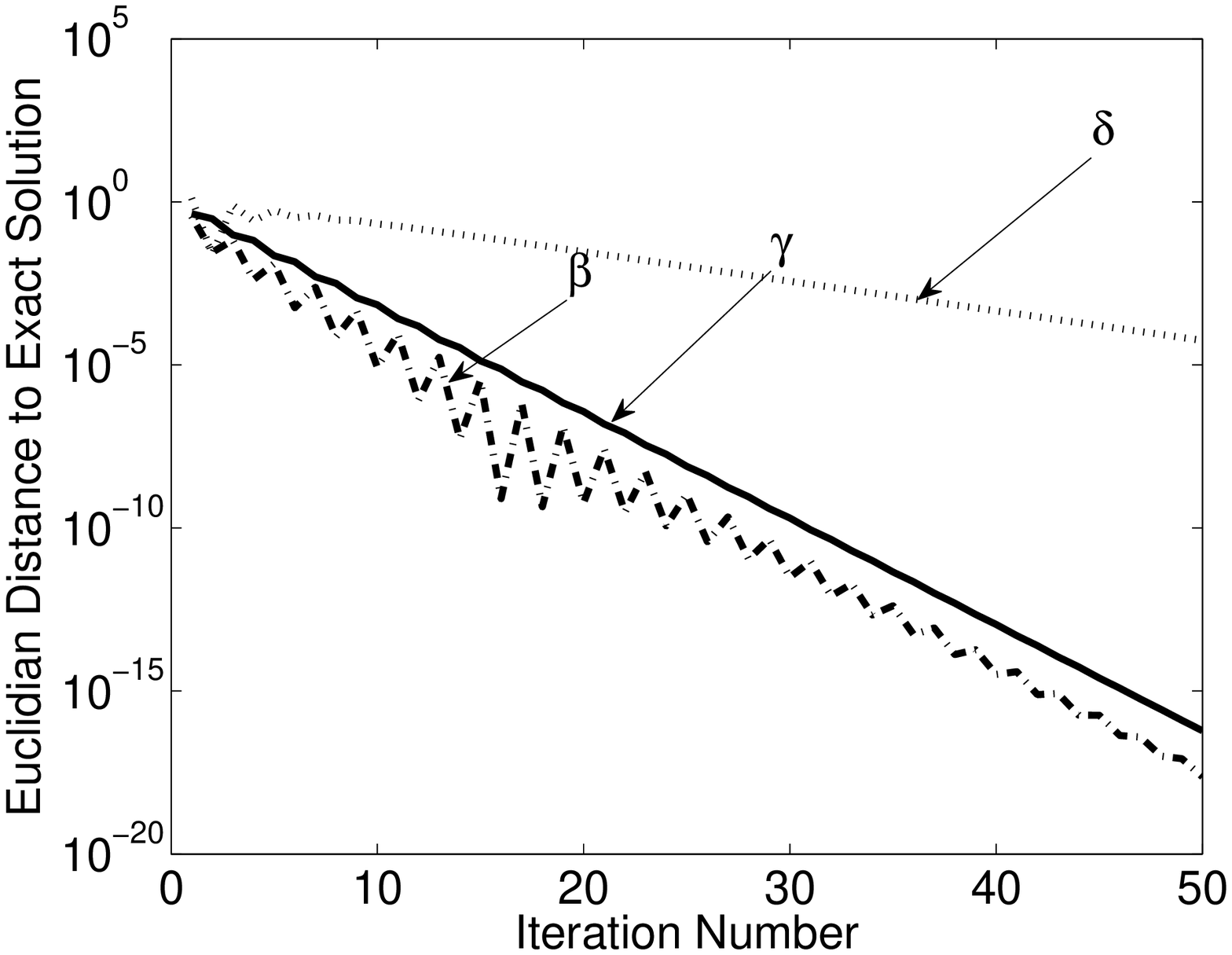}
    }
  \end{center}
    \caption{  Evolution dynamics of the Stable-Jacobi algorithm using CDMA correlation matrix $\mA_3$. All three parameters $\beta, \gamma, \delta$ are plotted per round, where the parameter of interest is the scale parameter $\delta$ (since i detection is performed by applying the $\sign$ operation on it. (a) Convergence of message norm (norm of the current posterior relative to  the posterior from previous round) when $\beta \ne 0$. (b) Distance of marginal to true solution (using regular plot).  (c)  Distance of the vector of marginals for the three users relative to the real transmission (using log plot).  (d) Marginal posterior for user 2.  The location parameter $\delta$ conforms to the binary transmission. (e) Distance between the posterior of user 2 to the true solution per iteration. }
\label{Jacobi-Evolution}
\end{figure*}

We demonstrate the properties Stable-Jacobi, using a small toy example.
Experimental settings are borrowed from \cite{Yener}. The linear transformation matrix is a synchronous CDMA channel transformation with a cross correlation matrix \small $\mA_3 = \frac{1}{7}\begin{pmatrix}7 & -1 & 3 \\
-1 & 7 & 5 \\
3 & -5 & 7 \\
\end{pmatrix}$\normalsize. As expected from the convergence analysis, the sufficient conditions for convergence hold since $\rho(|R_{3}|) = 0.9008 < 1$ and $\rho(|\mR_3|^{1.5}) = 0.6875<1$, and indeed the algorithm converges.
We initialized $x=[1,1,1]$ and the additive noise $Z_1 \sim \S(1.5,0,1,0),Z_2\sim \S(1.5,0.5,1,0), Z_3~\sim \S(1.5,0,1,0)$.
After computing $p(y)$, we computed $p(x| y)$ using Stable-Jacobi. Regarding convergence dynamics, convergence analysis shows that $\delta$ is converging more slowly since it is dependent on both $\beta$ and $\gamma$.
Figure \ref{Jacobi-Evolution}(a) shows convergence of message L2 norms. Figure \ref{Jacobi-Evolution}(b) plots the Euclidian distance of the intermediate solution on each round (as a vector in $\R^3$)\ to the exact solution computed by LCM-Stable. This distance indeed goes to zero as expected. Figure \ref{Jacobi-Evolution}(c) shows the same distance but using log plot. The almost straight line indicates that the distance is diminishing in a geometric fashion. Unlike the global distance which diminishes monotonically, when examining the second entry of the intermediate solution vector $x_2$ (Figure \ref{Jacobi-Evolution}(d)), we see the zigzag behavior which is a well known property of the Jacobi algorithm \cite{BibDB:BookBertsekasTsitsiklis}. This non-monotonic behavior is demonstrated also when  examining the Euclidian distance along the single dimension of $x_2$ (Figure \ref{Jacobi-Evolution}(e)).  

\subsection{Comparison to previous work}
In this section we compare our exact computation of inference in the linear stable model to previous techniques.  
Since stable distributions do not have a closed-form in the probability domain, any solution deployed in the pdf domain \textit{must involve approximation}. We investigate two existing approximate inference methods as a reference to our newly developed methods.  The first algorithm we implemented is Non-parametric belief propagation (NBP) \cite{NBP}. NBP works by approximating the observed stable distribution using a Gaussian mixture, and then computes a belief propagation procedure using a linear graphical model \cite{FaultDet}. The second algorithm  is Expectation Propagation (EP) variant of belief propagation \cite{EP}, which  approximates each Gaussian mixture message using a single Gaussian.

Algorithm 3 lists the approximation methodology we used.
Figure \ref{fig:EPNBP} depicts the different steps involved. The first tree steps prepare the input to the NBP/EP\ algorithms by converting the distribution into a mixture of Gaussians, with a relatively low 
number of mixture components (to allow for efficient execution). We construct a linear model graphical as described in \cite{FaultDet}. Then we run NBP/EP for a predefined number of rounds and output the computed belief. Next we can fit stable distribution parameter to the output. 

As well known, a drawback of the NBP algorithm is that the number of mixture components grows exponentially when computing the product step of the belief propagation algorithm. To avoid exponential blowup, efficient reduction methods where developed \cite{NBP2,briers05}. However, the efficiency comes at the cost of reduced accuracy. 

When working even with small problems (tens of variables) both algorithms did not perform well relative to our exact inference method. For example, on a 2D grid graph of 100 hidden nodes and 100 observation nodes, we got an average scale around 0.8146 while the average of true hidden scale was 1.  The averages of the skew and shift parameters where even worse, since they are dependent on the scale parameter.  

We tried to pinpoint to the root causes of reduced accuracy by constructing a small toy example. We constructed a small graphical model of two hodden nodes $X_1,X_2$ and two observed nodes $Y_1,Y_2$. The hidden node are initialized using a Cauchy distribution with variance 1. To further simplify we set all the edge weights to one, so $\mA = \begin{pmatrix}1 &1 \\ 1& 1\end{pmatrix}$. Observations are received using the linear transformation $\vy = \mA\vx$.

  Even for this small problem we can see (Fig. \ref{fig:EPNBP}(d)) that the NBP output does not match exactly the true solution.
We believe that the largest error is rooted in the product 
step approximation. 

Another possible approach is to use Expectation Propagation. EP\ operates by approximating each Gaussian mixture with a single mixture, creating a light weight and faster approximation (relative to NBP).
Fig. \ref{fig:EPNBP}(e) shows an EP approximation of  a single mixture. For Cauchy distributions, EP captured quite well the shape of the distribution (Fig. \ref{fig:EPNBP}(f)), but less well the exact mean. However, for skewed distributions, EP does not capture well the distribution shape, since the distribution shape can not be approximated using a single Gaussian (Fig. \ref{fig:EPNBP}(g)). 
\begin{figure}[ht!]
\centering{
\fbox{
\begin{minipage}{5.6in}\small
\setlength{\baselineskip}{5mm} 
\scriptsize
{\tt \begin{enumerate}\item Quantisize the stable distribution.
\item Fit a Gaussian mixture to the quantisized observation using Kernel Ridge Regression.
\item Optional: reduce the number of mixture components using sampling techniques.
\item Run non-parametric belief propagation  \cite{NBP} or expectation propagation \cite{EP}.
\item Quantisize the resulting mixture.
\item Fit a stable distribution to the quantization and retrieve the parameters.
\end{enumerate}} 
\end{minipage} 
}
}\\
\vspace{2mm}
Algorithm 3: Approximate inference for linear stable model.\ 
\end{figure}
\begin{figure}[h!]
  \begin{center}
    \subfigure[Step 1: quantization]{
      \includegraphics[width=.22\textwidth]{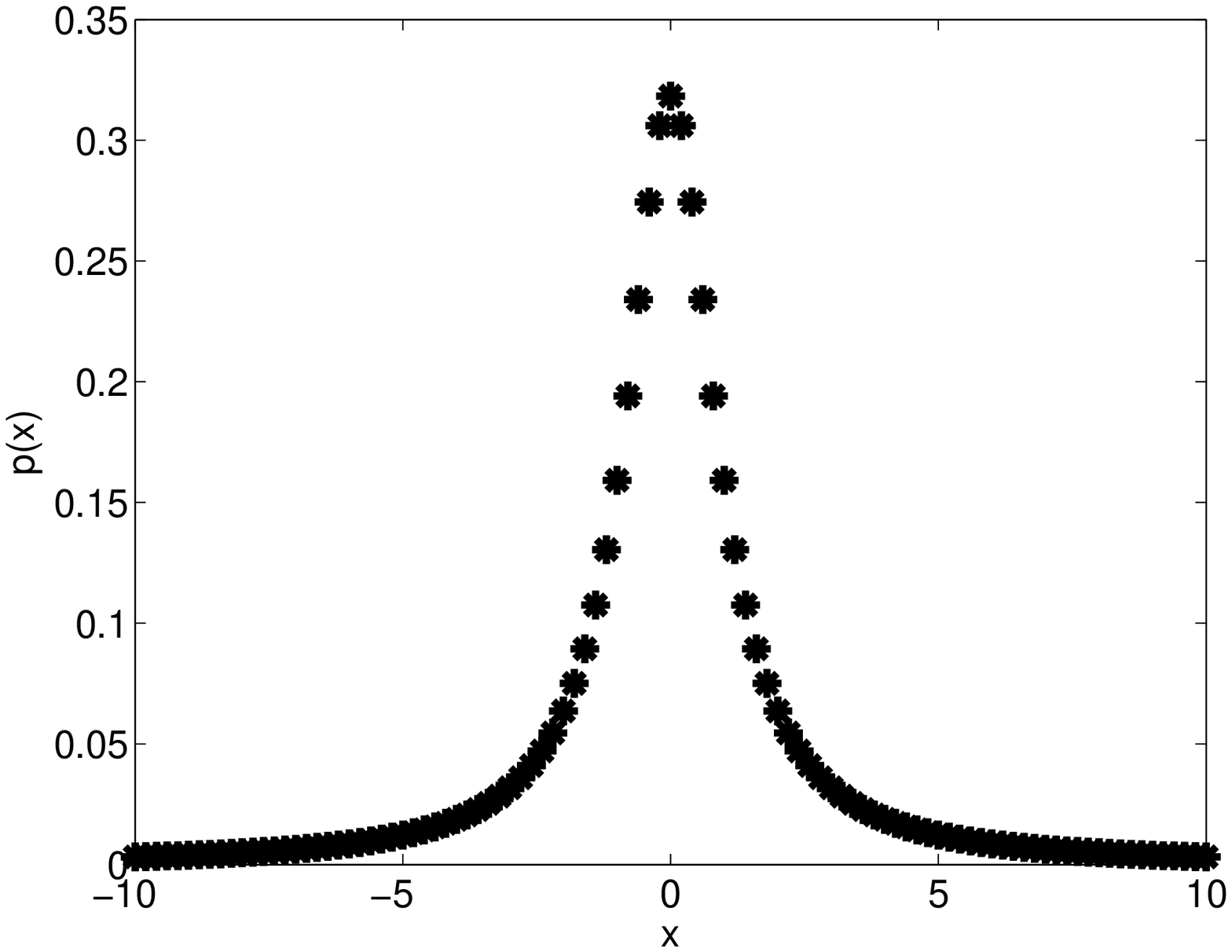}
    }
     \subfigure[Step 2: fitting a Gaussian mixture]{
      \label{fig:CoemScalabilityPlota}
      \includegraphics[width=.22\textwidth]{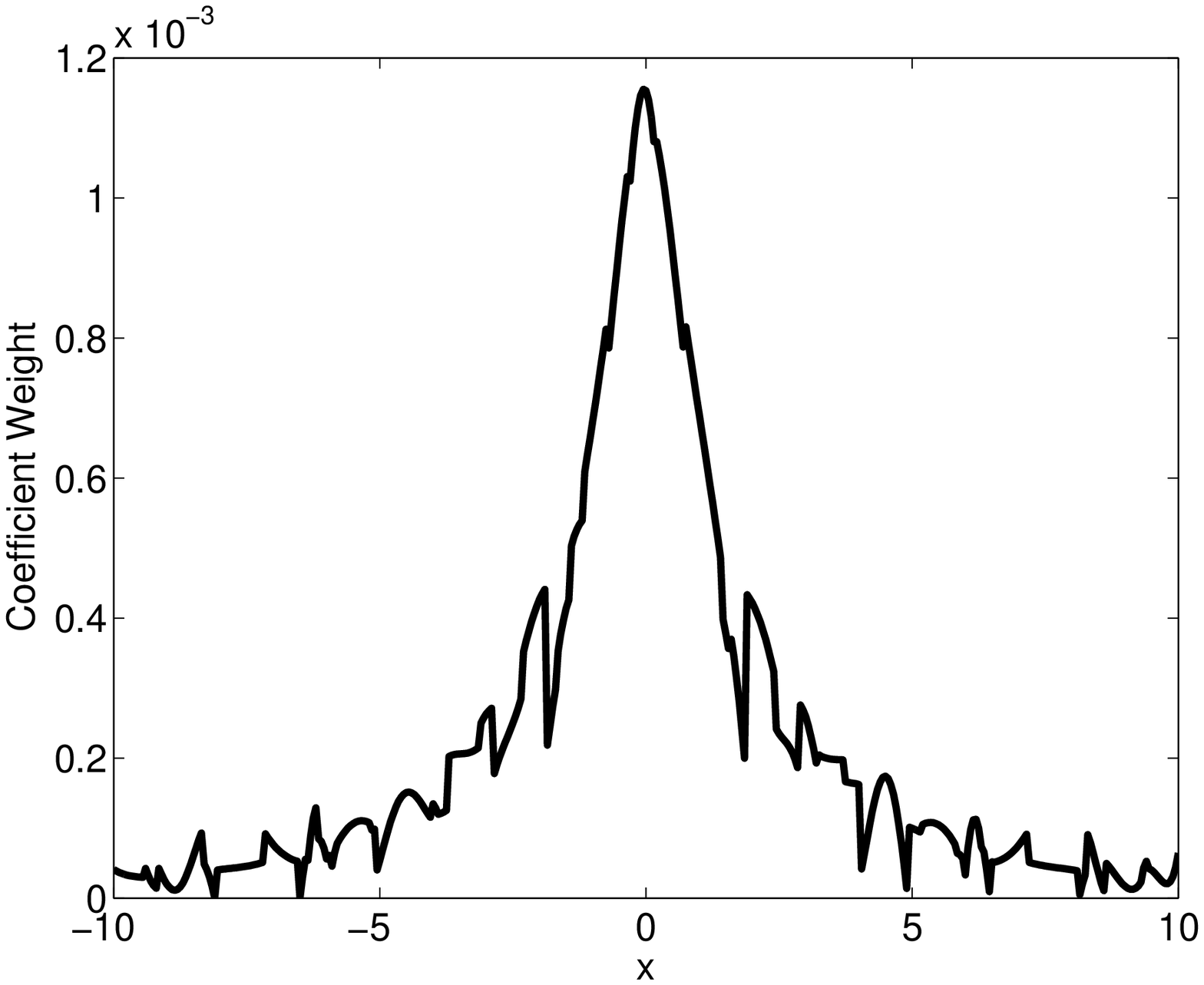}
    }
      \subfigure[Step 3: Resampling]{
      \includegraphics[width=.245\textwidth]{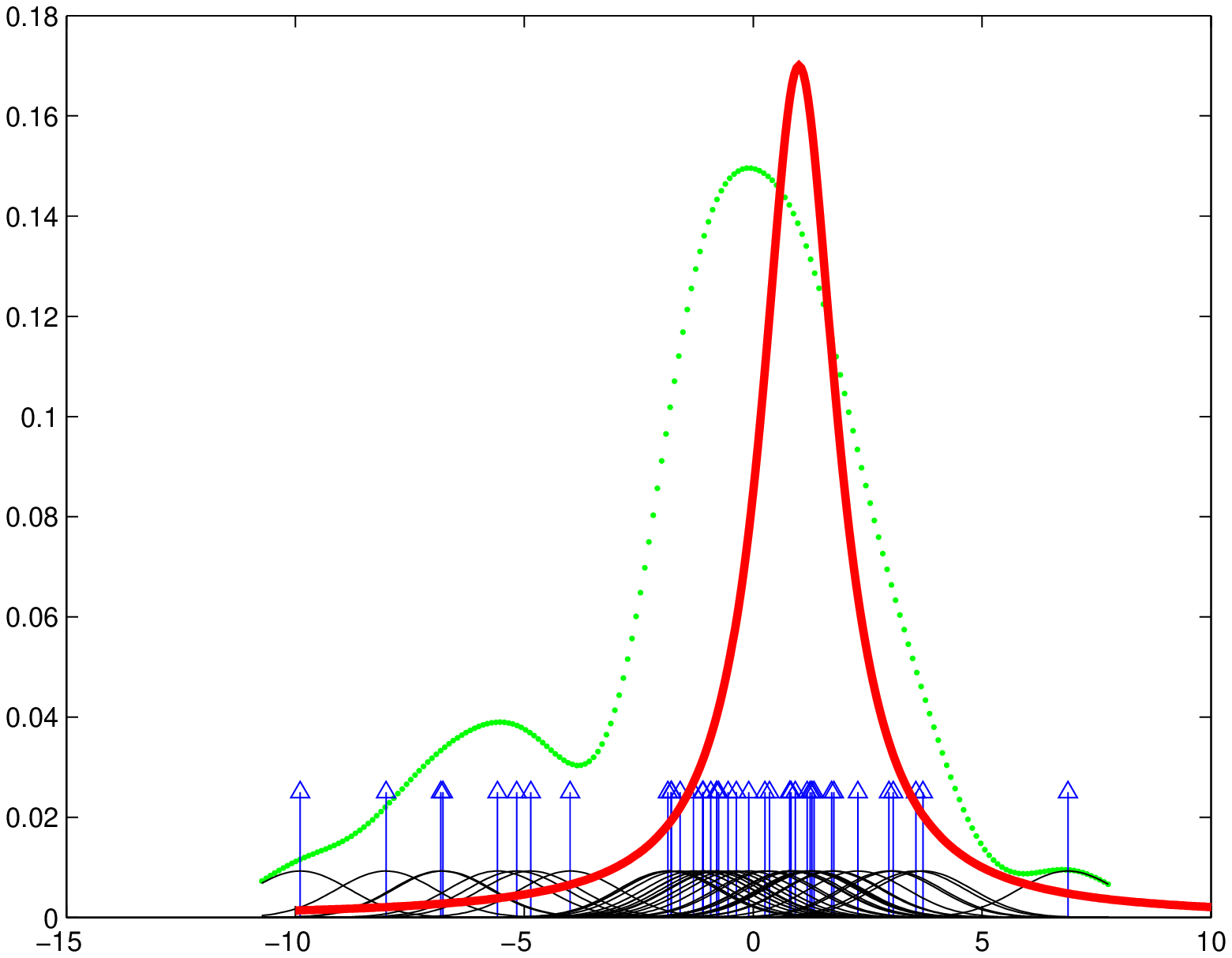}
    }
     \subfigure[Step 4: Output of NBP]{
      \includegraphics[width=.22\textwidth]{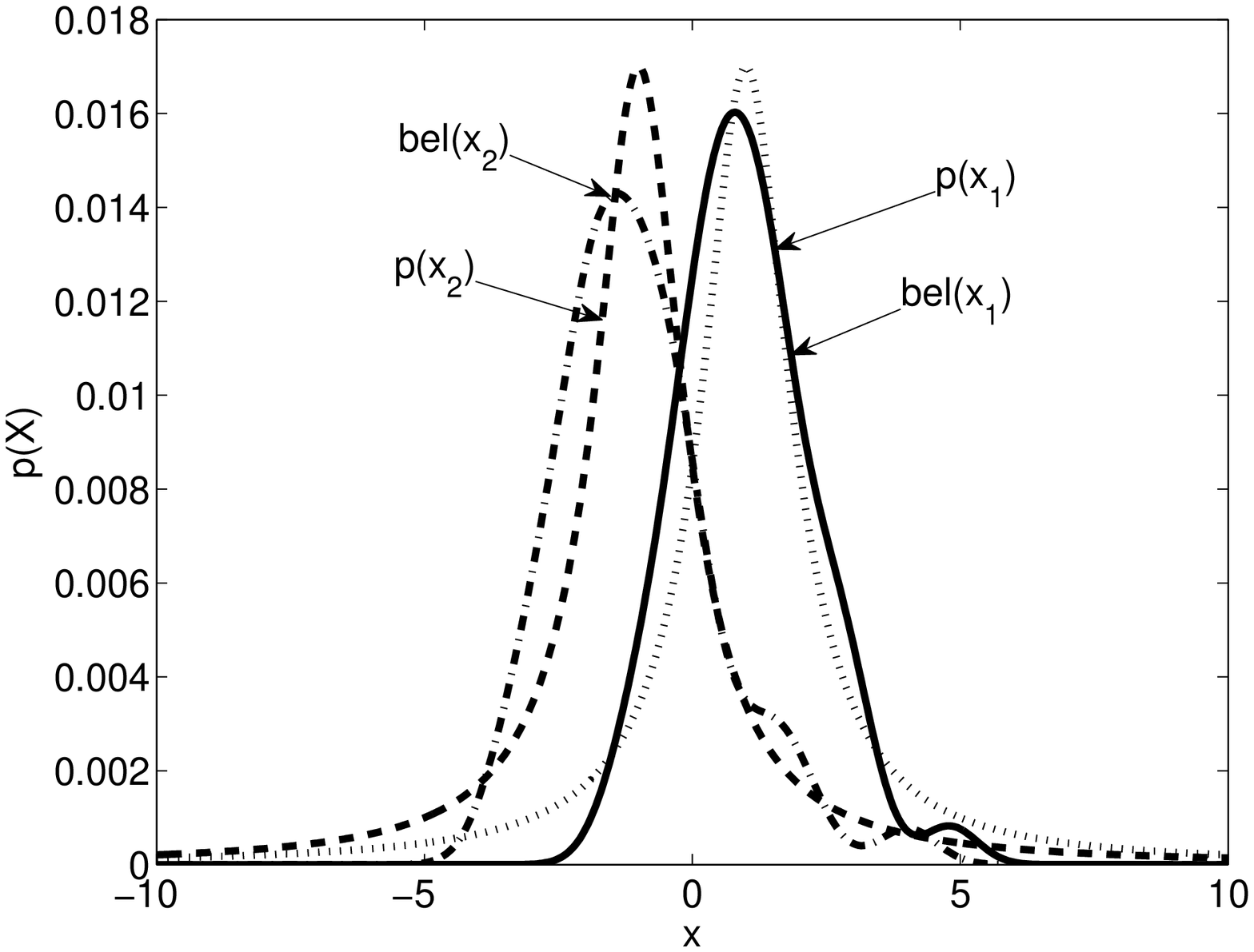}
    }
     \subfigure[Expectation-Propagation estimate]{
      \includegraphics[width=.26\textwidth]{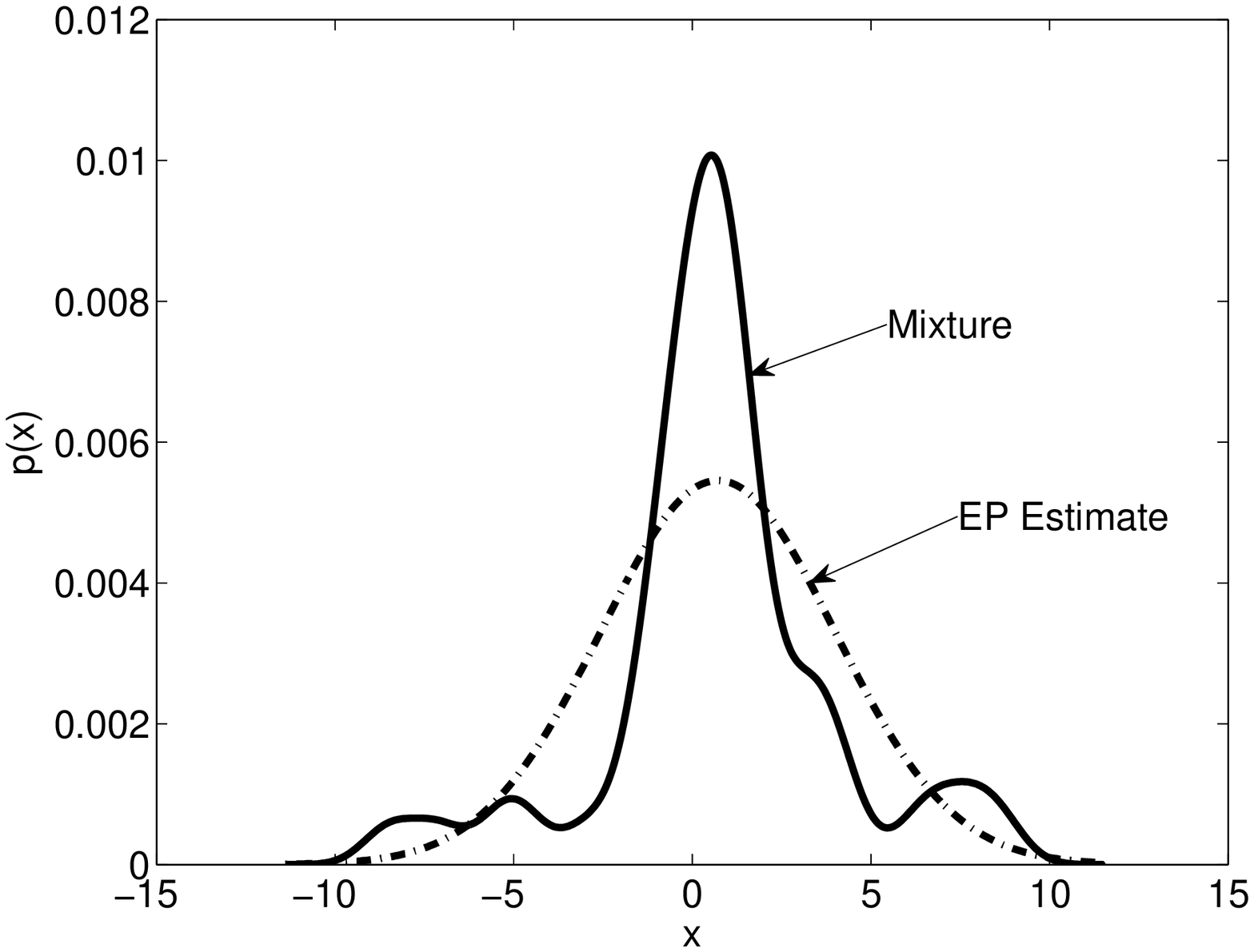}
    }
     \subfigure[Step 4: output of EP (Cauchy prior)]{
      \includegraphics[width=.26\textwidth]{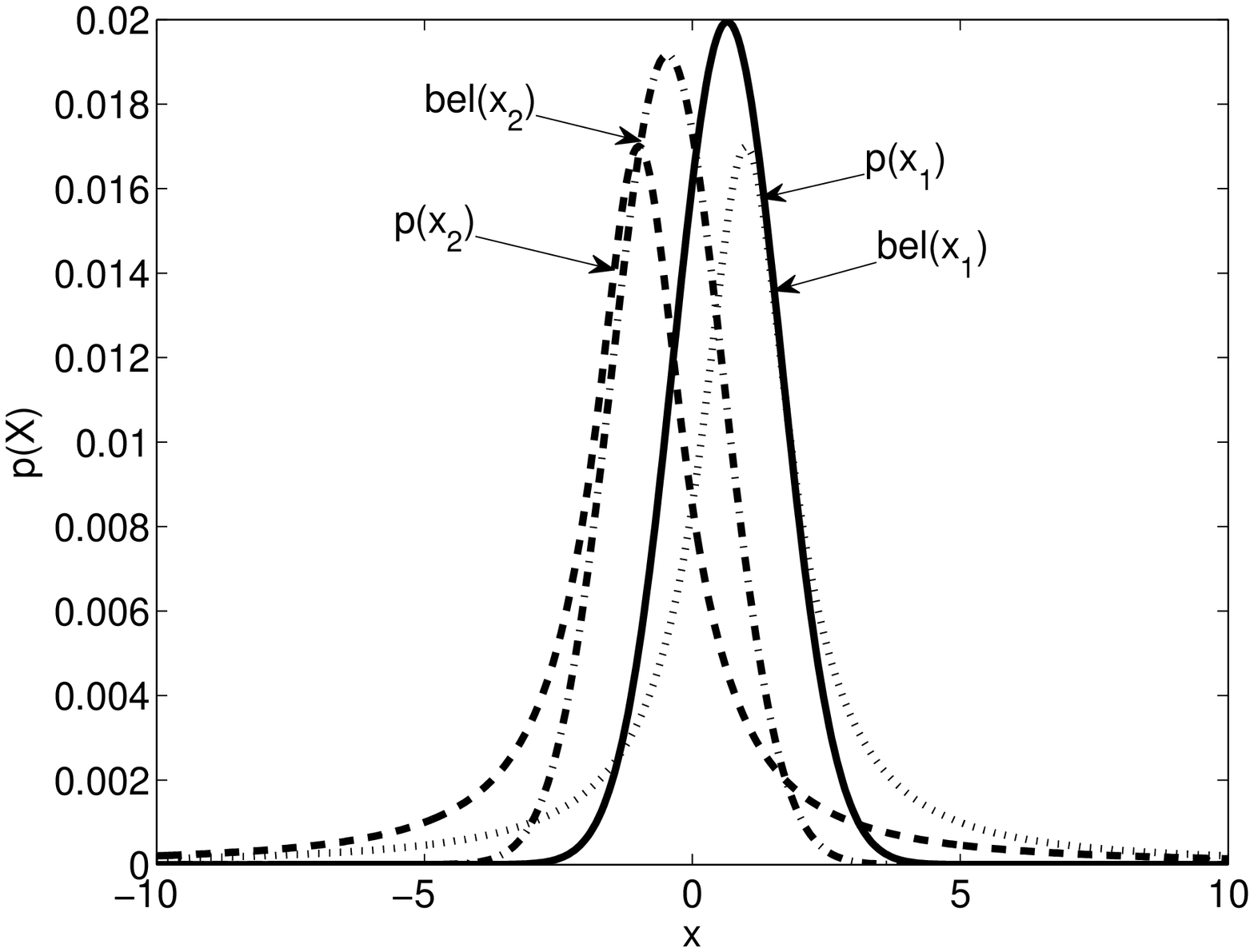}
    }
     \subfigure[Step 4: output of EP (L\'evy)\ prior]{
      \includegraphics[width=.26\textwidth]{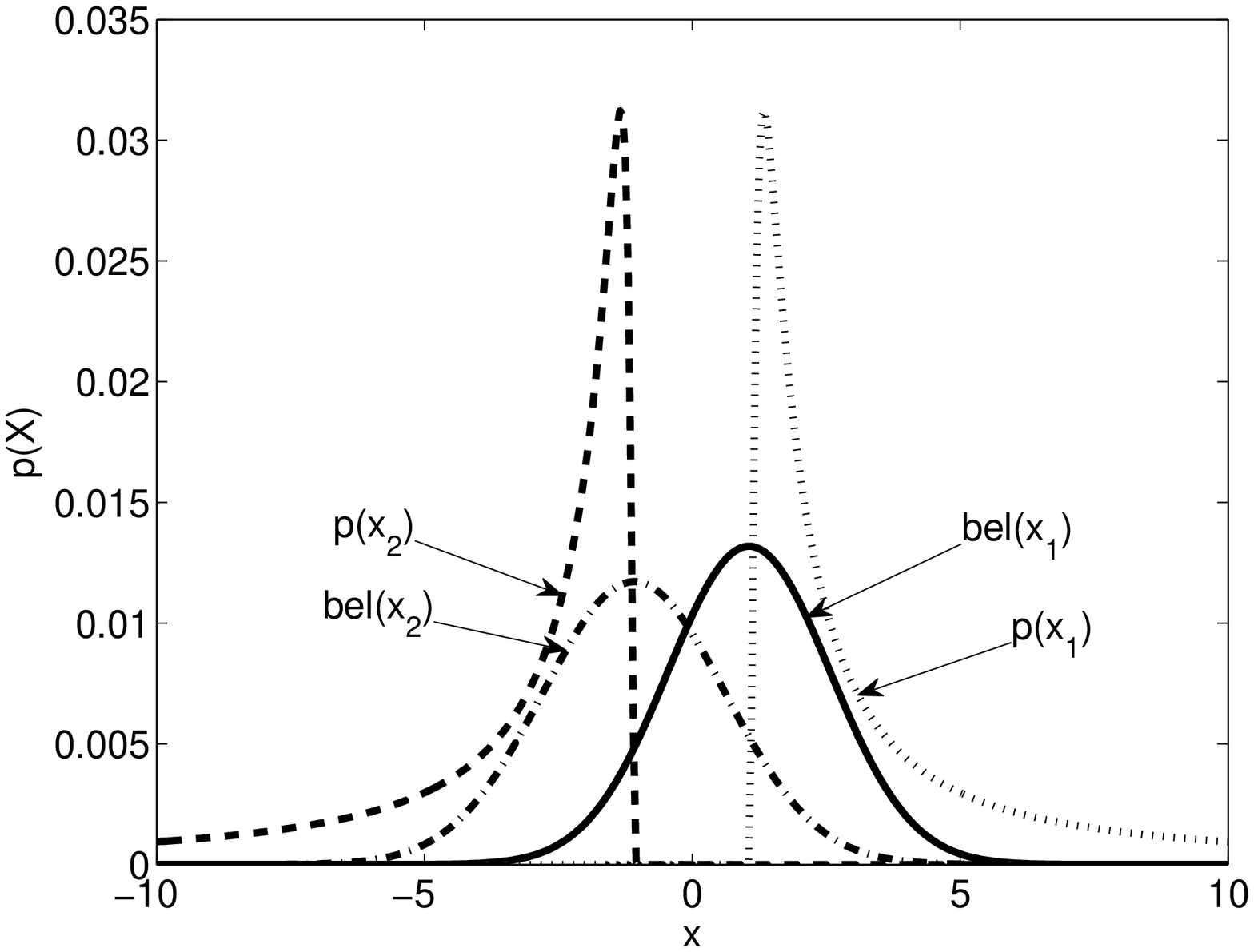}
    }
   \end{center}
   \caption{Approximated inference using previous techniques: non-parametric BP and Expectation Propagation. In contrast, Stable-Exact computes inference directly in this model by inverting 3 matrices.}
\label{fig:EPNBP}
\end{figure}

Overall, we conclude that using previous techniques, it is significantly more difficult to compute inference in a linear-stable model and the results obtained are not accurate. In contrast, using our developed exact inference procedure the solution is obtained exactly by simply computing three matrix inverses. 

\end{document}